%% file: main.tex
\documentclass[10pt,twocolumn,letterpaper]{article}

\usepackage[pagenumbers]{cvpr} %
\def\confName{CVPR}
\def\confYear{2027}

\makeatletter
\g@addto@macro\caption@subtypehook{\let\ltx@label\label}
\makeatother

\newif\ifdebug
\debugfalse

\usepackage[utf8]{inputenc} %
\usepackage[T1]{fontenc}    %
\usepackage{microtype}      %

\usepackage{nicefrac}       %
\usepackage{arydshln}
\expandafter\let\csname endtabular*\endcsname\endtabular

\usepackage{algorithm}
\usepackage[noend]{algpseudocode} %
\algrenewcommand\algorithmicrequire{\textbf{Input:}}
\algrenewcommand\algorithmicensure{\textbf{Output:}}

\input{config/commands.tex}
\input{config/acronyms.tex}

\definecolor{cvprblue}{rgb}{0.21,0.49,0.74}
\usepackage[pagebackref,breaklinks,colorlinks,allcolors=cvprblue]{hyperref}

\title{Star Panoptic Quality: Selectively Evaluating Split and Merged Instances}
\input{config/authors.tex}

\begin{document}

\maketitle

\input{body/0_abstract.tex}

\acresetall

\input{body/1_introduction.tex}
\input{body/2_related_work.tex}

\input{body/3_method.tex}
\input{body/4_experiments.tex}

\input{body/5_conclusion.tex}
\input{body/6_llm_usage_statement.tex}

\input{body/7_acknowledgements}

{\small
\bibliographystyle{ieeenat_fullname}
\bibliography{references}
}

\clearpage
\setcounter{page}{1}
\appendix
\input{body/8_appendix.tex}

\ifdebug
  \clearpage
  \input{config/rules.tex}
\fi

\end{document}

%% file: config/acronyms.tex
\usepackage[nolist]{acronym}

\newacro{TP}{true positive}
\newacro{FP}{false positive}
\newacro{FN}{false negative}

\newacro{IoU}{Intersection over Union}
\newacro{IoL}{Intersection over Leaf}
\newacro{CC}{connected components}

\newacro{OS}{oversegmentation}
\newacro{US}{undersegmentation}

\newacro{1:1}{One-to-One}
\newacro{M:1}{Many-to-One}
\newacro{1:M}{One-to-Many}
\newacro{M:M}{Many-to-Many}

\newacro{PQ}{Panoptic Quality}
\newacro{RQ}{Recognition Quality}
\newacro{SQ}{Segmentation Quality}
\newacro{PS}{Panoptic Segmentation}
\newacro{StarPQ}{Star Panoptic Quality}

\newacroindefinite{FP}{an}{a}
\newacroindefinite{FN}{an}{a}
\newacroindefinite{ROI}{an}{a}

%% file: config/authors.tex
\newcommand{\authorWithAffiliation}[2]{\textbf{#1}\textsuperscript{#2}}

\author{%
  \authorWithAffiliation{Erik Großkopf}{1},
  \authorWithAffiliation{Paula Tamara Buzduga}{1},
  \authorWithAffiliation{Hendrik Möller}{2,3},
  \authorWithAffiliation{Florian Kofler}{1,4,5,6}\\[1ex]
  \small\textsuperscript{1}Hertie Institute for AI in Brain Health, University of Tübingen, Tübingen, Germany\\
  \small\textsuperscript{2}Department of Neuroradiology, TUM University Hospital, Munich, Germany\\
  \small\textsuperscript{3}Chair for AI in Healthcare and Medicine, Technical University of Munich (TUM), Munich, Germany\\
  \small\textsuperscript{4}AI for Image-Guided Diagnosis and Therapy, TUM School of Medicine and Health,\\
  \small Technical University of Munich, Munich, Germany\\
  \small\textsuperscript{5}Department of Quantitative Biomedicine, University of Zurich, Zurich, Switzerland\\
  \small\textsuperscript{6}Helmholtz AI, Helmholtz Munich, Neuherberg, Germany\\
  \small\texttt{erik.grosskopf@uni-tuebingen.de}%
}

%% file: body/0_abstract.tex
\begin{abstract}
Detecting individual instances is crucial for many segmentation problems.
However, expert instance annotations are often unavailable, so instances are approximated heuristically (e.g., as connected components of semantic maps), which can merge touching objects or split fragmented ones.
Even when expert annotations exist, expert disagreement or structural delineation biases can yield split or merged reference instances, corresponding to over- or under-segmentation, respectively. 
Panoptic Quality (PQ) and related metrics capture both segmentation and detection quality while weighting all instances equally, regardless of size.
Yet when the references are biased, these metrics can neither detect split or merged instances nor distinguish valid from invalid over- or under-segmentation.
We present Star Panoptic Quality (StarPQ), a novel evaluation framework, which closes this gap in two steps: 
First, we construct a so-called star forest in the bipartite graph whose clean, merge, and split stars isolate different failure modes.
Second, we calculate the final metric by individually weighting each failure mode, allowing practitioners to account for instance delineation biases. 
Across various experiments, we demonstrate that StarPQ not only enables targeted penalization or leniency towards specific failure modes, but also serves as a diagnostic tool to reveal their prevalence.
\end{abstract}

\input{figures/teaser/teaser}

%% file: figures/teaser/teaser.tex
\begin{figure}
\captionsetup[subfigure]{aboveskip=0mm,belowskip=.4mm}
\begin{subfigure}{0.495\linewidth}
 \includegraphics[width=1.0\linewidth,trim=0 1cm 0 4cm,clip]{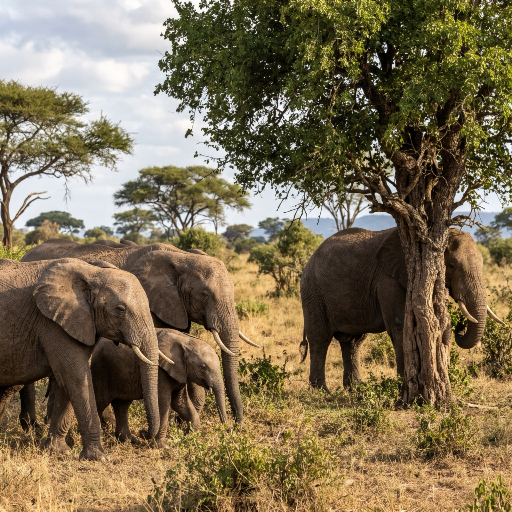}
 \subcaption{image}\label{fig:image}
\end{subfigure}
\begin{subfigure}{0.495\linewidth}
 \includegraphics[width=1.0\linewidth,trim=0 1cm 0 4cm,clip]{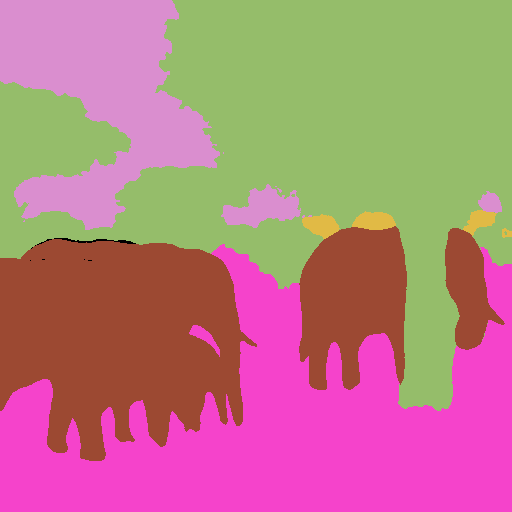}
 \subcaption{semantic map}\label{fig:semantic}
\end{subfigure}\\
\begin{subfigure}{0.495\linewidth}
 \includegraphics[width=1.0\linewidth,trim=0 1cm 0 4cm,clip]{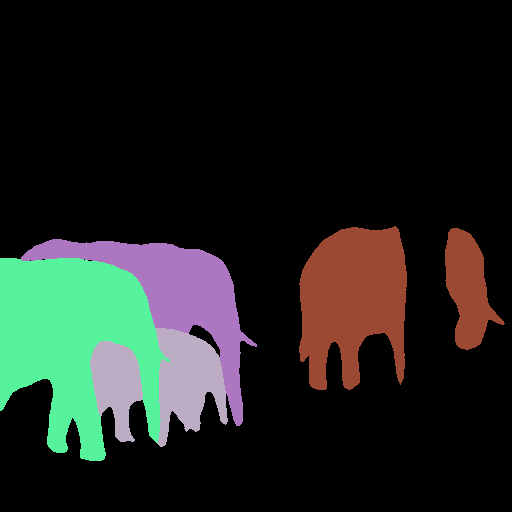}
 \subcaption{instance map}\label{fig:instance}
\end{subfigure}
\begin{subfigure}{0.495\linewidth}
 \includegraphics[width=1.0\linewidth,trim=0 1cm 0 4cm,clip]{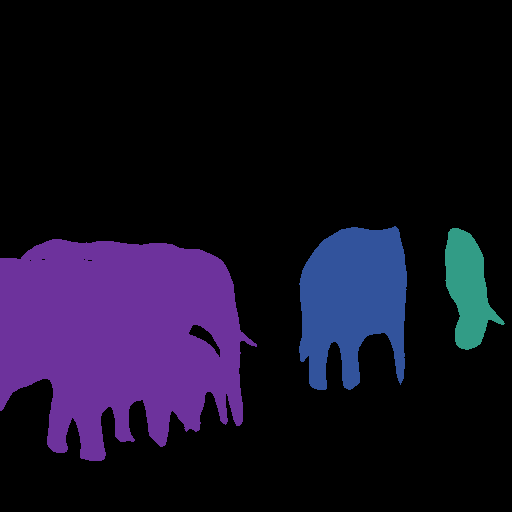}
 \subcaption{connected components}\label{fig:cc}
\end{subfigure}
\caption{
For a given image \subref{fig:image}, we show its semantic map \subref{fig:semantic} and instance map \subref{fig:instance}. In many segmentation datasets, instance maps do not exist.
Hence, practitioners resort to recovering instance maps from semantic maps, e.g., via connected components \subref{fig:cc}, a process that is error-prone.
In this example, mutually occluding elephants in the left herd merge into a single instance (purple), while partial occlusion by the tree wrongfully splits the right elephant into two instances (blue and green).
Thus, not a single elephant is recovered correctly.
}
\label{fig:examples}
\end{figure}

%% file: body/1_introduction.tex
\input{figures/bipartite_graph/bipartite_graphs}

\section{Introduction} \label{sec:introduction}
In many segmentation problems, the detection and accurate delineation of individual instances are of great importance.
Examples include detecting metastases or multiple sclerosis lesions in the medical domain~\citep{kofler2023blob}, quantifying different cell types in microscopy datasets~\citep{sigmund2023genetically}, or detecting individual cars in autonomous driving~\citep{cordts2016cityscapes}.
\ac{PS} unifies the tasks of semantic and instance segmentation, and model performance in this task may be evaluated by \ac{PQ}~\citep{kirillov2019panoptic}, where each reference segment contributes to the score with equal weight, irrespective of its size. 
Computing \ac{PQ} therefore requires instance-level reference labels.
Although many semantic segmentation tasks may be viewed as instance segmentation problems, segmentation datasets often lack instance labels due to the additional labor and costs of the annotation process.
Practitioners recover instances from semantic maps using methods such as \ac{CC} or \textit{watershed} in order to train and evaluate instance segmentation models on such datasets (\cref{fig:examples}).
In many datasets, the recovered instances are only approximations, because the assumptions of the recovery method and the dataset's acquisition method may introduce biases in instance delineation.

How a dataset was acquired determines which causes of splits and merges can arise when instances are recovered from semantic maps.
Because natural photographs project a 3D scene onto the image plane, objects may occlude one another, in which case the connected components of a semantic map can differ from the true instances: a partially occluded instance can be split into several components (an elephant behind a tree), while adjacent or mutually occluding instances can merge into one (animals in a herd).
In volumetric data such as CT or MRI, the split failure due to occlusion does not occur because instances cannot be hidden behind anything. 
The merge failure, however, remains: touching structures still merge into a single instance, as is the case with adjacent cells in 3D microscopy or tightly packed organs.
In all rasterized imagery, limited resolution can merge neighboring instances whenever the gap between them falls below the acquisition's effective resolution.
We speak of \ac{OS} when several predicted instances describe a single reference instance, and of \ac{US} when a single predicted instance describes several reference instances.
Depending on the task, dataset acquisition, instance recovery, or downstream requirements, penalization or leniency towards individual failure modes can be required.

\ac{PQ} matches predictions and references \ac{1:1} at a fixed \ac{IoU} threshold of 0.5, the lowest threshold that guarantees a unique matching.
This \ac{1:1} matching penalizes correctly predicted instances that were merged or recovered as multiple split fragments in the approximated instance map.
Existing alternatives either retain a \ac{1:1} matching~\citep{karmakar2025softpq,chen2023sortedap} or do not measure both failure modes symmetrically~\citep{grosskopf2026redefining}.
To address these limitations, we present \ac{StarPQ}, a novel instance segmentation metric that quantifies the occurrence of splits and merges and allows practitioners to penalize or forgive each failure mode individually.
\ac{StarPQ}'s assignment, the \textit{star forest}, is constructed from the \ac{IoL} rather than the \ac{IoU}. 
Unlike an \ac{IoU} threshold of 0.5, the \ac{IoL} criterion allows a single segment to be described by multiple segments, and we apply it in both directions, so that a reference can be described by several predictions (a split) and a prediction by several references (a merge).
The \textit{star forest} is a set of \textit{stars} that share no segments, each consisting of a center segment and one or more leaf segments from the other map.
Every matched segment therefore belongs to exactly one star: a clean star with a single leaf, or a split or merge star with several.

In our experiments we show how prior information about dataset acquisition and instance recovery can guide the configuration of the metric, preventing the penalization of correct predictions and accommodating downstream-specific requirements.
In addition, \ac{StarPQ} can serve as a diagnostic tool during training or instance recovery, providing insights into the occurrence and abundance of splits and merges.
Finally, we propose and demonstrate a novel protocol that allows incorporating expert disagreement about instance delineation into an evaluation pipeline.
Our contributions are i) a star forest construction that detects splits and merges without double counting segments, ii) the \ac{StarPQ} metric, which decomposes \acp{TP} into clean, split, and merge stars, and iii) a protocol to model expert disagreement about instance delineation.

%% file: figures/bipartite_graph/bipartite_graphs.tex
\begin{figure*}[t]
    \centering
    \newcommand*{\rowHeight}{5cm}%
    \newcommand*{\separator}{\hfill\vrule\hfill}%
    \subcaptionbox{Overlapping maps\label{fig:bipartite_graph:overlapping_maps}}{%
        \hspace{1em}%
        \includegraphics[height=\rowHeight]{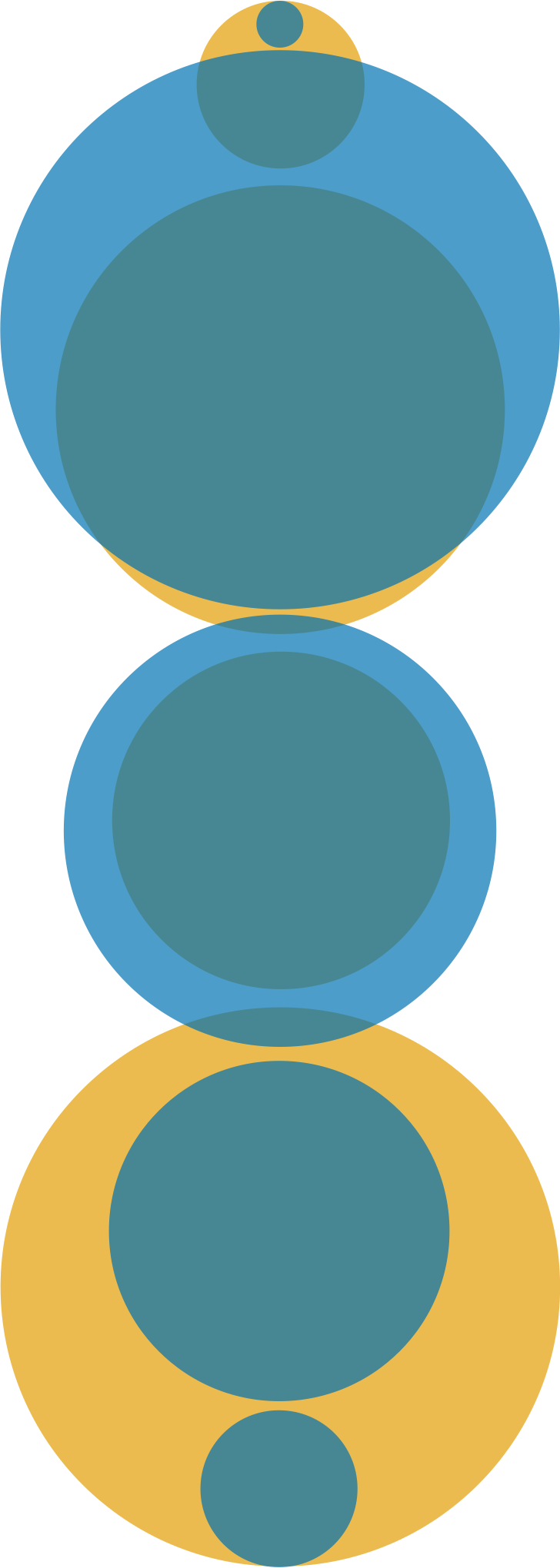}%
        \hspace{1em}}%
    \separator
    \subcaptionbox{IoU overlap graph\label{fig:bipartite_graph:overlap_graph}}{%
        \includegraphics[height=\rowHeight]{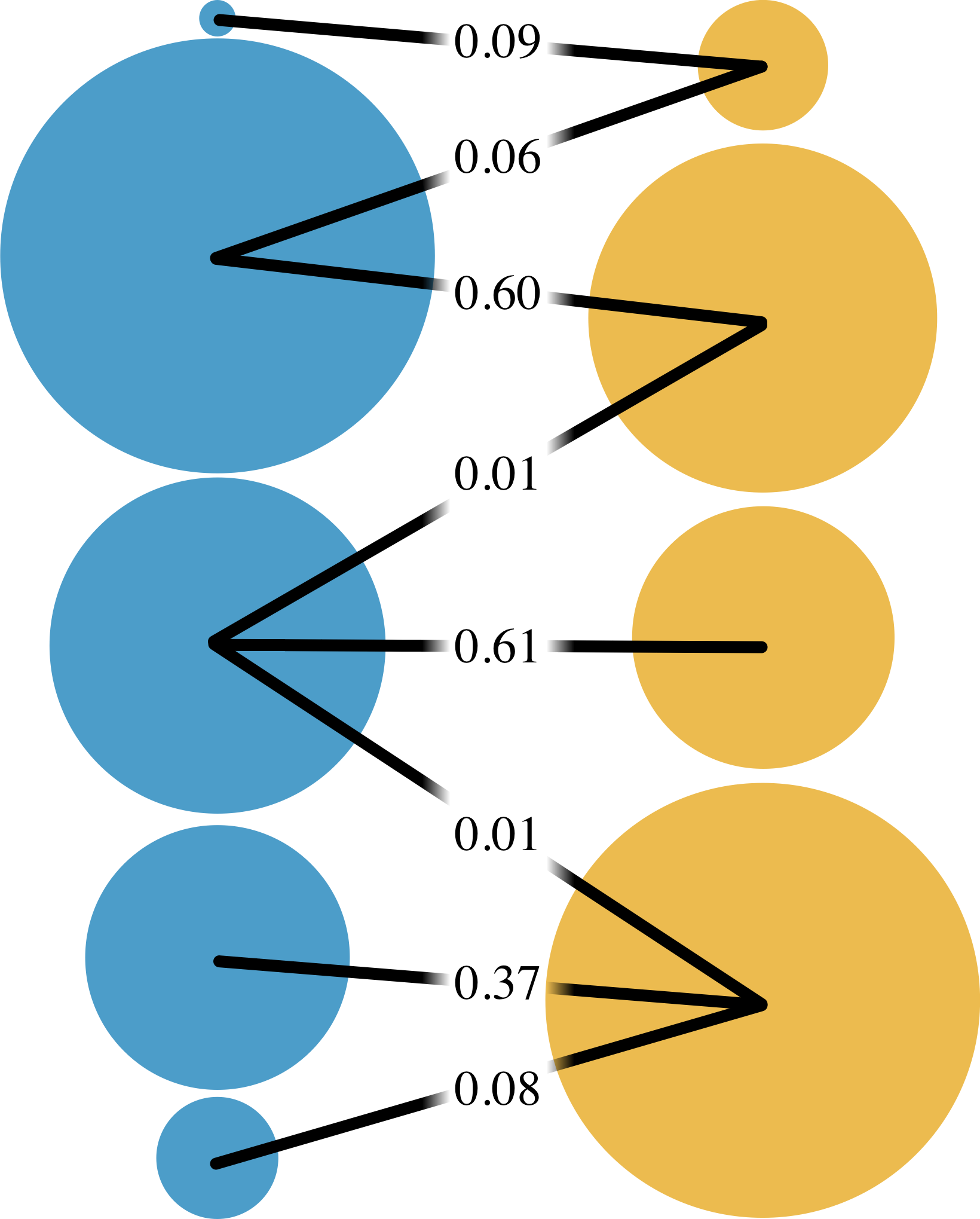}}%
    \separator
    \subcaptionbox{One-to-One matching (PQ)\label{fig:bipartite_graph:pq_matching}}{%
        \includegraphics[height=\rowHeight]{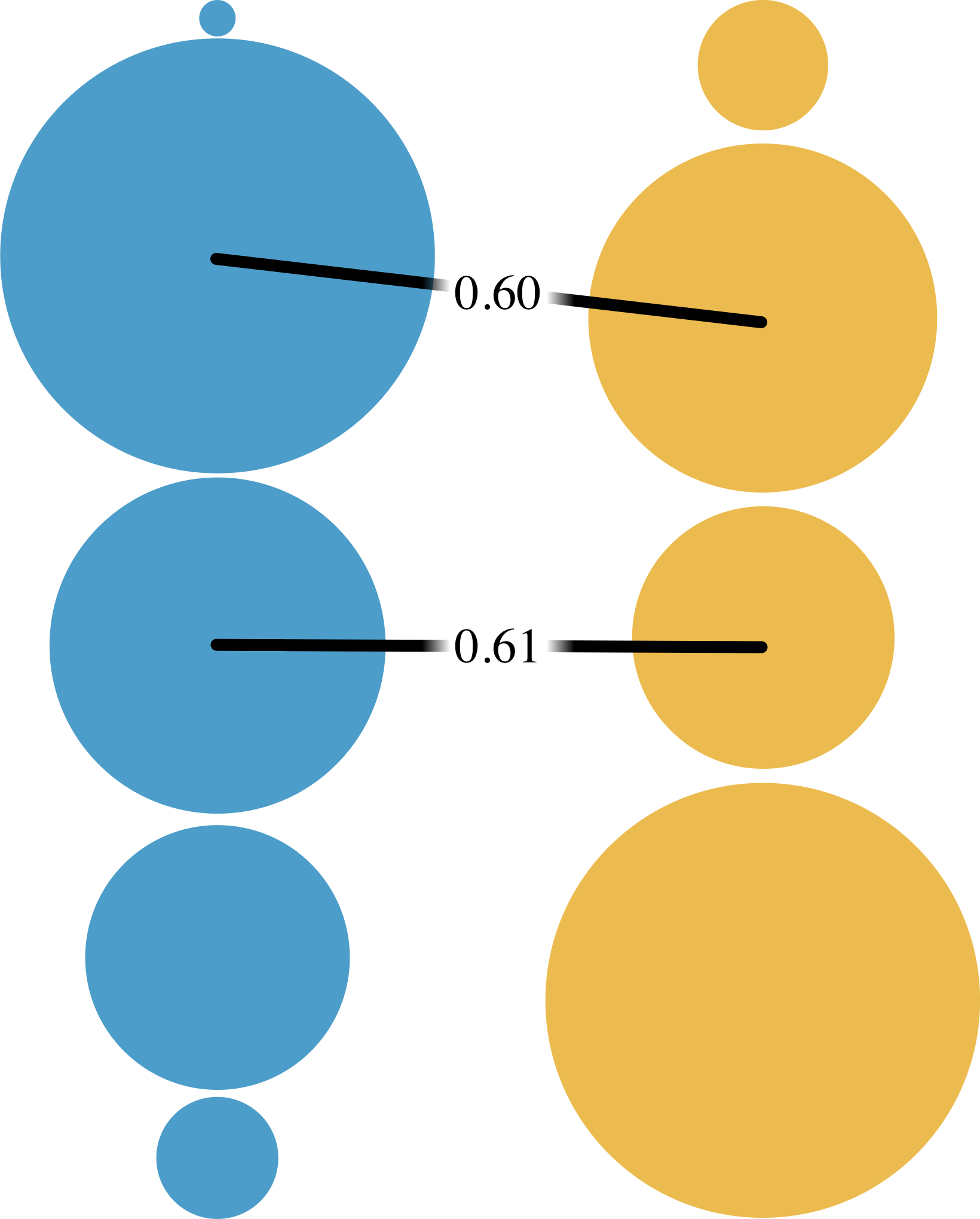}}%
    \separator
    \subcaptionbox{Star Forest (StarPQ)\label{fig:bipartite_graph:star_forest}}{%
        \includegraphics[height=\rowHeight]{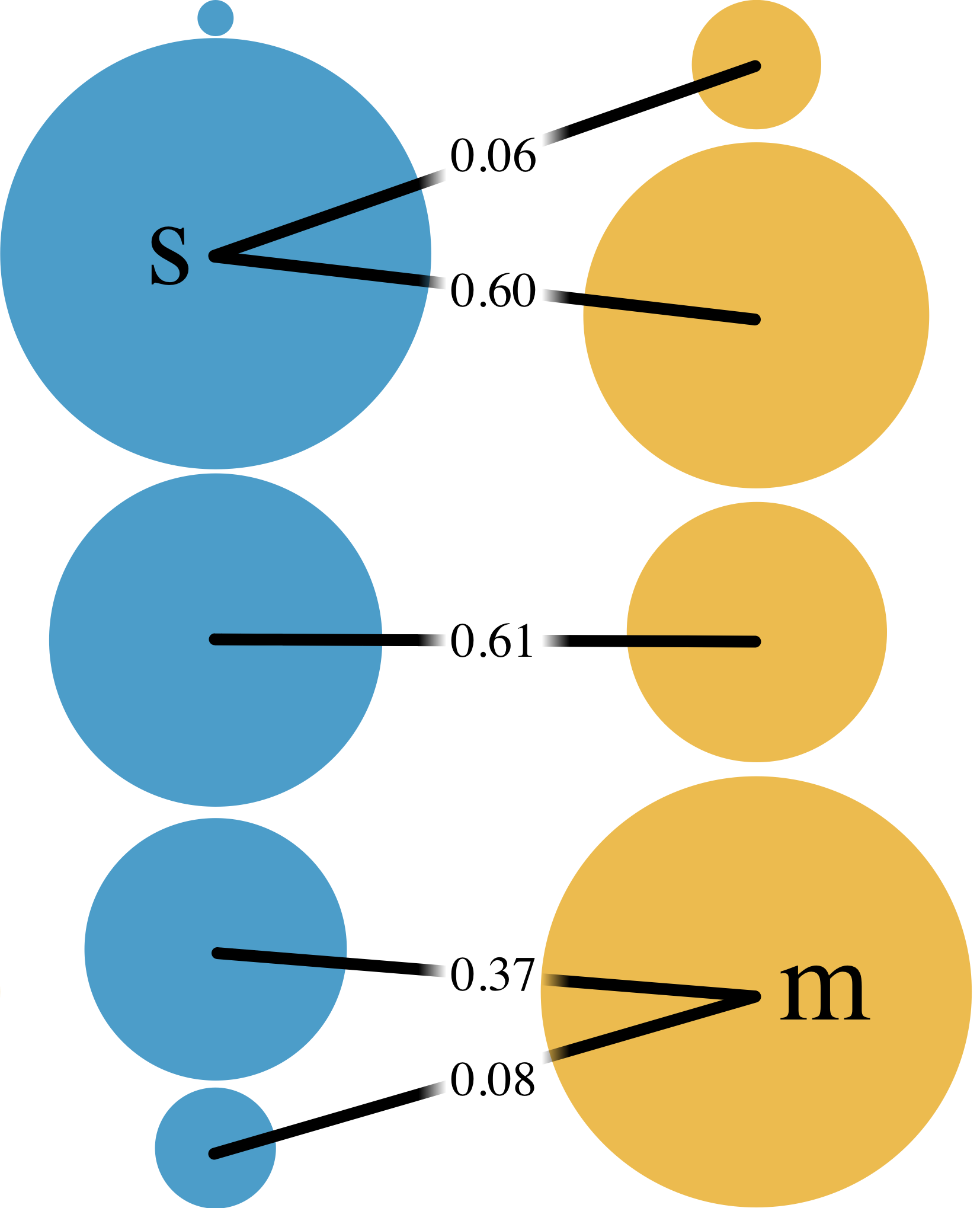}}
    \caption{For the given reference (blue) and prediction (yellow) maps \subref{fig:bipartite_graph:overlapping_maps}, we show the bipartite overlap graph \subref{fig:bipartite_graph:overlap_graph} with edges corresponding to overlapping pairs of reference and prediction instances and IoU as edge weight (predictions are offset horizontally for readability of edges).
    Panoptic Quality's matching \subref{fig:bipartite_graph:pq_matching} selects all edges with $\IoU>0.5$, yielding a one-to-one assignment, whereas StarPQ's star forest construction \subref{fig:bipartite_graph:star_forest} permits assignment of many predictions to one reference (split star, top), one prediction to one reference (clean star, center), and many references to one prediction (merge star, bottom).
    The reference instance s is the \textit{center} of a split star with two \textit{leaf} prediction instances.
    The prediction instance m is the \textit{center} of a merge star with two \textit{leaf} reference instances.
    }
    \label{fig:bipartite_graph}
\end{figure*}

%% file: body/2_related_work.tex
\section{Related Work} \label{sec:related_work}
In 2019, \citet{kirillov2019panoptic} introduced \ac{PS}, which unifies semantic and instance segmentation into a single coherent task.
To each pixel of a reference, a semantic class (\textit{stuff} or \textit{thing}) and, for \textit{thing} classes, an instance ID are assigned, effectively asking models to predict both a semantic and an instance map.
By allowing at most one instance per pixel, overlapping reference instances are not possible by definition.
To measure model performance, \citet{kirillov2019panoptic} propose \ac{PQ}, which decomposes into a \ac{SQ} and a \ac{RQ} term:
\begin{equation}\label{eq:pq}
\PQ
= \underbrace{\frac{\sum_{(p,g) \in \TP} \IoU(p,g)}{|\TP|}}_{\SQ}
\times \underbrace{\frac{|\TP|}{|\TP| + \tfrac{1}{2}|\FP| + \tfrac{1}{2}|\FN|}}_{\RQ},
\end{equation}
where $\TP$ is the set of matched pairs $(p, g)$ of a predicted segment $p$ and a reference segment $g$ with \ac{IoU} defined by $\IoU(p,g) = |p \cap g| / |p \cup g|$ that is strictly greater than 0.5.
The \ac{SQ} term measures the average \ac{IoU} of all overlapping pairs that satisfy this constraint.
$\FP$ and $\FN$ denote the predicted and reference segments, respectively, that remain unmatched from this \ac{1:1} assignment (\cref{fig:bipartite_graph:pq_matching}).
Note that each pair in $\TP$ contributes equally, regardless of its size, as \ac{IoU} measures overlap relatively.
Further, this criterion yields a \acl{1:1} matching which is both optimal and unique as depicted in \cref{fig:bipartite_graph:pq_matching}. 
Within their guidance framework, \textit{Metrics Reloaded}, \citet{maier-hein2024metrics} recommend \ac{PQ} for instance segmentation problems, as it assesses segmentation quality and detection performance with a single score.
\citet{chen2023sortedap} propose sortedAP, which bypasses a fixed threshold by scoring at all unique thresholds of all matched pairs and integrating over the threshold.
\citet{karmakar2025softpq} introduce SoftPQ, aiming to soften the hard threshold of \ac{PQ} by instead employing two thresholds, where the lower one acts as a more lenient one.
\citet{kofler2023approaching} introduce the concept of \textit{Peak Ground Truth}, where model improvement beyond a certain threshold may reproduce erroneous annotations.

\paragraph{Bipartite formulation.}
In the context of \ac{PS}, \citet{chazalon2021revisiting} formalized the assignment within a bipartite graph, with reference and prediction instances corresponding to nodes and overlapping pairs corresponding to edges.
From the \ac{PS} non-overlap property, it follows that there cannot be edges within the sets of references or predictions, allowing us to split nodes into two disjoint sets.
This formulation yields a weighted bipartite overlap graph, in which the edge weight may be defined by the \ac{IoU} (\cref{fig:bipartite_graph:overlap_graph}).

\paragraph{Beyond One-to-One matchings.}
\ac{PQ}'s 0.5 matching threshold guarantees a \ac{1:1} matching that is both optimal and unique.
\citet{grosskopf2026redefining} explored different matching strategies that emerge when dropping the lower bound of 0.5 on the threshold.
They propose to constrain the degree of reference or prediction nodes within the bipartite graph, yielding four matching strategies. 
In \ac{1:1}, each prediction matches at most one reference, and vice versa, penalizing \ac{OS} and \ac{US} symmetrically.
\ac{M:1} matching allows many predictions to jointly describe a single reference by constraining only prediction node degrees, representing leniency towards \ac{OS}.
\ac{1:M} matching works the other way around and is lenient towards \ac{US}, but may count prediction segments multiple times and dilute \ac{SQ}.
Finally, they argue that \ac{M:M} falls outside this matching taxonomy, as it relates sets of references to sets of predictions, effectively clustering instances rather than constructing a meaningful and interpretable assignment.
\ac{PQ}~\citep{kirillov2019panoptic} supports \textit{crowd} annotations for image regions in which many instances of the same class are densely packed and individual instance masks are either unavailable or ambiguous, effectively allowing many predictions to describe a single region and modeling leniency in such situations.
To mitigate the bias towards larger connected components, \citet{jaus2025every} propose \ac{CC}-Metrics, which partition the image into Voronoi regions around each reference component, so predictions crossing a boundary are split rather than assigned as a whole.
To evaluate cell segmentation in videos as part of the Cell Tracking Challenge, \citet{maska2014benchmark} matched a reference and a segmented cell if their intersection was greater than half the area of the reference cell.
This yields at most one prediction per reference cell, and possibly multiple references per prediction, corresponding to a \ac{1:M} matching.

%% file: body/3_method.tex
\section{Star Panoptic Quality} \label{sec:method}
In this section, we introduce a novel metric, called \ac{StarPQ}, that symmetrically quantifies the failure modes \ac{OS} and \ac{US}.
\ac{StarPQ} gives practitioners fine-grained control to penalize or forgive each failure mode individually.
Previous metrics or variations only describe or penalize either split- or merge-components, but none is capable of measuring both occurrences symmetrically without double counting instances.
We aim to extend \ac{PQ}'s desideratum of interpretability~\citep{kirillov2019panoptic} not only by decomposing into segmentation and recognition quality terms, but also by decomposing \acp{TP} into three distinct categories that isolate specific characteristics.
Further, we believe that simplicity of matching and decomposition is important, as the assignment should be easy and fast to compute.
Like \ac{PQ}, \ac{StarPQ} counts each segment at most once, evaluates each class independently and may be averaged across all classes.
Unlike \ac{PQ}, whose \ac{IoU} threshold of 0.5 is what makes its matching unique, \ac{StarPQ} obtains uniqueness from a \textit{leaf criterion} (\cref{eq:leaf_criterion}), so its star threshold $\tau$ can be chosen freely.
\ac{StarPQ} aims to measure model performance with finer granularity than \ac{PQ} by leveraging these properties.

\subsection{Star Forests}
\ac{PQ} matches reference and predicted segments \acl{1:1}.
In contrast, we procedurally construct a \emph{star forest} in the bipartite overlap graph, whose nodes are the reference and predicted segments, and whose edges connect overlapping pairs (\cref{fig:bipartite_graph:star_forest}).
A \emph{star} consists of one \emph{center} and one or more \emph{leaves} on the opposite side, each connected only to this particular center.
A \textit{star forest} is a set of pairwise node-disjoint stars.
Stars with exactly one leaf are \emph{clean stars}, in which the center and leaf are interchangeable (center pair in \cref{fig:bipartite_graph:star_forest}).
Those with two or more leaves are \emph{split stars} if centered on a reference segment, and \emph{merge stars} if centered on a predicted segment (top and bottom components in \cref{fig:bipartite_graph:star_forest}).
Thus, every segment is either unmatched or belongs to exactly one clean, split, or merge star.
For a star $s$ with reference segments $G_s$ and predicted segments $P_s$, we define $\IoU(s) = \IoU\bigl(\bigcup G_s, \bigcup P_s\bigr)$, treating each side as a single segment.

\input{figures/star_forest/candidate_graph}

\subsection{Star Forest Construction}
A segment $l$ points to a segment $c$ of the other map, written $l \rightarrow c$, if $c$ holds the majority of its area:
\begin{equation}\label{eq:leaf_criterion}
    |l \cap c| > \tfrac{1}{2}|l|.
\end{equation}
We call this the \textit{leaf criterion}.
Equivalently, the \acl{IoL}, $\IoL(l,c) = |l \cap c| \,/\, |l|$, is strictly greater than $0.5$.
By the non-overlap property of \ac{PS}, at most one segment can hold the majority of $l$, so every segment points to at most one other, and these pointers form the candidate graph $C$ (\cref{fig:candidate_graph}).
If no segment were both pointing and being pointed to, $C$ would already be a star forest:
The segments pointed to would be the centers, and those pointing to them their leaves.
The leaf criterion, however, only constrains the leaf. 
A segment that holds the majority of others can still give the majority of its own area to another segment, and is then both a center and a leaf.
This happens in two ways: two segments can point to each other, forming a \emph{mutual pair} $g \leftrightarrows p$, which is typical for a good match, or pointers can form chains such as $g_1 \rightarrow p_1 \rightarrow g_2$.
We therefore keep a single star in each group of segments connected by pointers, centered where its pointers end: at the segment that points nowhere or, for a mutual pair, at the one whose star, with all segments pointing to it, has the higher \ac{IoU}.
Its leaves are the segments pointing directly to it, and all other segments of the group remain unmatched. 
In the chain above, where $g_2$ points nowhere, $p_1$ thus becomes a leaf of $g_2$, and $g_1$ is unmatched.
Every group has exactly one such end. 
A pointer carries more than half of its segment and is therefore that segment's largest overlap, so overlaps strictly increase along a chain, $|g_1 \cap p_1| < |p_1 \cap g_2|$, and following the pointers never leads in a circle, except between the two segments of a mutual pair.
We compare this procedural construction with the selection that maximizes the sum of star \acp{IoU}, computed exactly via dynamic programming. 
We find that chains are rare in practice, and our heuristic approach coincides with the optimal solution in the vast majority of cases (in COCO, the two differ in 0.1--0.2\,\% of samples).
Finally, a star threshold $\tau$ removes every star with $\IoU(s) \le \tau$ and leaves its segments unmatched.
The construction itself does not depend on the metric's configuration.

\input{figures/overlap_shift/shift_score_figure}

\subsection{StarPQ Computation}
The \acp{TP} are the stars of the star forest, and the \acp{FP} and \acp{FN} are the predicted and reference segments left unmatched.
Each star counts as a single \ac{TP}, so a split star has $|P_s| - 1$ extra prediction segments and a merge star $|G_s| - 1$ extra reference segments:
\begin{equation}
  \FP_{\text{split}} = \sum_{s \in \TP} \bigl(|P_s| - 1\bigr), \qquad
  \FN_{\text{merge}} = \sum_{s \in \TP} \bigl(|G_s| - 1\bigr).
\end{equation}
The weighted counts $\FP^{w}$ and $\FN^{w}$ charge unmatched segments in full and extra segments by $\lambda_{\text{split}}, \lambda_{\text{merge}} \in \{0, 1\}$:
\begin{equation}
\begin{aligned}
  \FP^{w} &= |\FP| + \lambda_{\text{split}}\,\FP_{\text{split}}, \\
  \FN^{w} &= |\FN| + \lambda_{\text{merge}}\,\FN_{\text{merge}}.
\end{aligned}
\end{equation}
$\lambda_{\text{split}}$ and $\lambda_{\text{merge}}$ act as switches to individually penalize extra split or merge segments when enabled.
At $\lambda_{\text{split}} = \lambda_{\text{merge}} = 1$, every segment is charged exactly once, as in \ac{PQ}. 
When both switches are disabled, the extra segments of splits or merges are forgiven entirely.
Since $\lambda_{\text{split}}$ and $\lambda_{\text{merge}}$ are either 0 or 1, the weighted counts are integers corresponding to either only unmatched segments or unmatched and extra segments.
\ac{StarPQ} is defined as
\begin{equation}\label{eq:pcq}
  \StarPQ
  = \underbrace{\frac{\sum_{s \in \TP} \IoU(s)}{|\TP|}}_{\SQ}
    \times
    \underbrace{\frac{|\TP|}{|\TP| + \tfrac{1}{2}\FP^{w}
      + \tfrac{1}{2}\FN^{w}}}_{\RQ},
\end{equation}
which has the structure of \ac{PQ}, with stars in place of matched pairs.
Our definition achieves our goal of uniformly and symmetrically measuring split and merge components by decomposing \ac{TP} into the respective categories.
The unique star forest construction, which yields disjoint stars, preserves interpretability by attributing each segment exactly once.

\subsection{Comparison to PQ and Existing Metrics}
The leaf criterion is the matching rule used in the Cell Tracking Challenge~\citep{maska2014benchmark}, which applies it only to reference segments.
In \ac{StarPQ}, its symmetric application to both reference and predicted segments yields the candidate graph.
The \ac{1:1} matching of \ac{PQ} is generally more restrictive than the construction of the star forest of \ac{StarPQ}, which additionally allows \ac{1:M} and \ac{M:1} components.
SoftPQ and sortedAP also rely on a \ac{1:1} matching.
The framework of \citet{grosskopf2026redefining} formalizes isolated \ac{M:1} or \ac{1:M} matchings which exclusively express leniency towards \ac{OS} or \ac{US}, respectively. 
In contrast, \ac{StarPQ} measures these failure modes symmetrically without double counting segments.
Note that \ac{StarPQ} is not upper-bounded by the global \ac{IoU}: \ac{IoU} weighs each component by its size, while \ac{StarPQ} weighs every component equally.
Hence, \ac{StarPQ} can exceed \ac{IoU} when small instances are delineated more accurately than big ones.

%% file: figures/star_forest/candidate_graph.tex
\begin{figure}[tb]
    \centering
    \includegraphics[height=5cm]{figures/bipartite_graph/bipartite_overlap.png}%
    \hspace{1em}\vrule\hspace{1em}%
    \includegraphics[height=5cm]{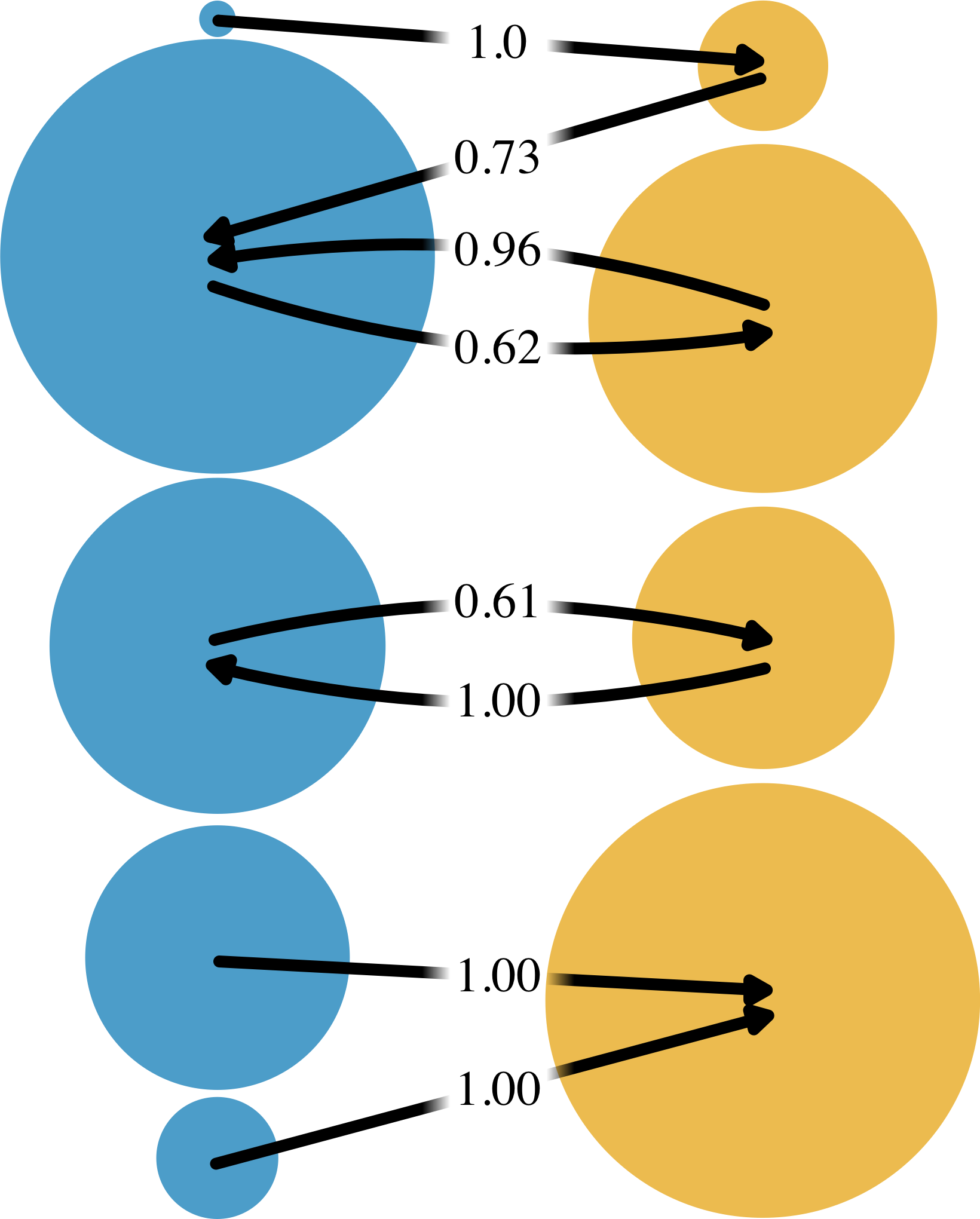}
    \caption{
    Overlapping reference (blue) and prediction (yellow) maps (left) and candidate graph $C$ (right) of the example in \cref{fig:bipartite_graph}.
    An arrow $l \rightarrow c$ is drawn wherever the intersection over leaf is greater than 0.5 and is labeled with that value.
    The top component is a chain ending in a mutual pair, the middle component is a mutual pair, and the bottom component is already a star, with two leaves on the left pointing to one center on the right.
    }
    \label{fig:candidate_graph}
\end{figure}

%% file: figures/overlap_shift/shift_score_figure.tex
\begin{figure*}[t]
    \centering
    \resizebox{\textwidth}{!}{%
        \includegraphics[height=4cm]{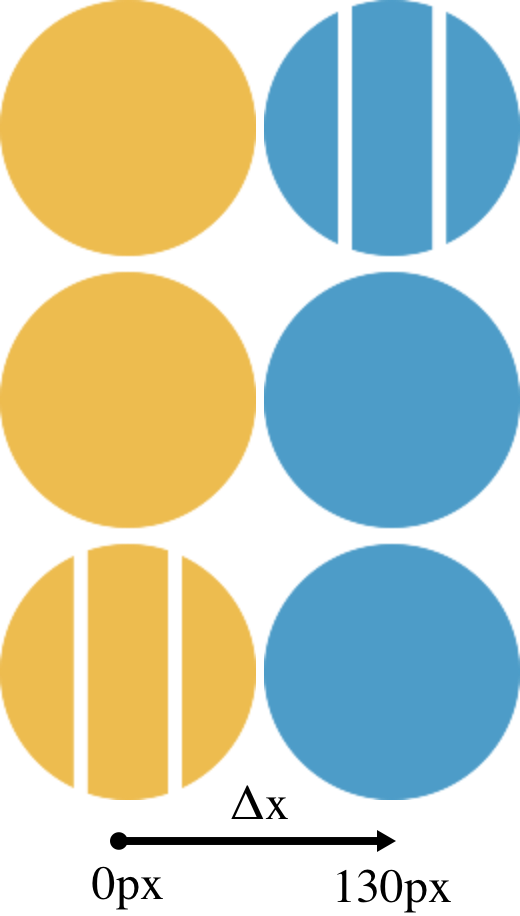}%
        \hspace{1em}%
        \includegraphics[height=4cm]{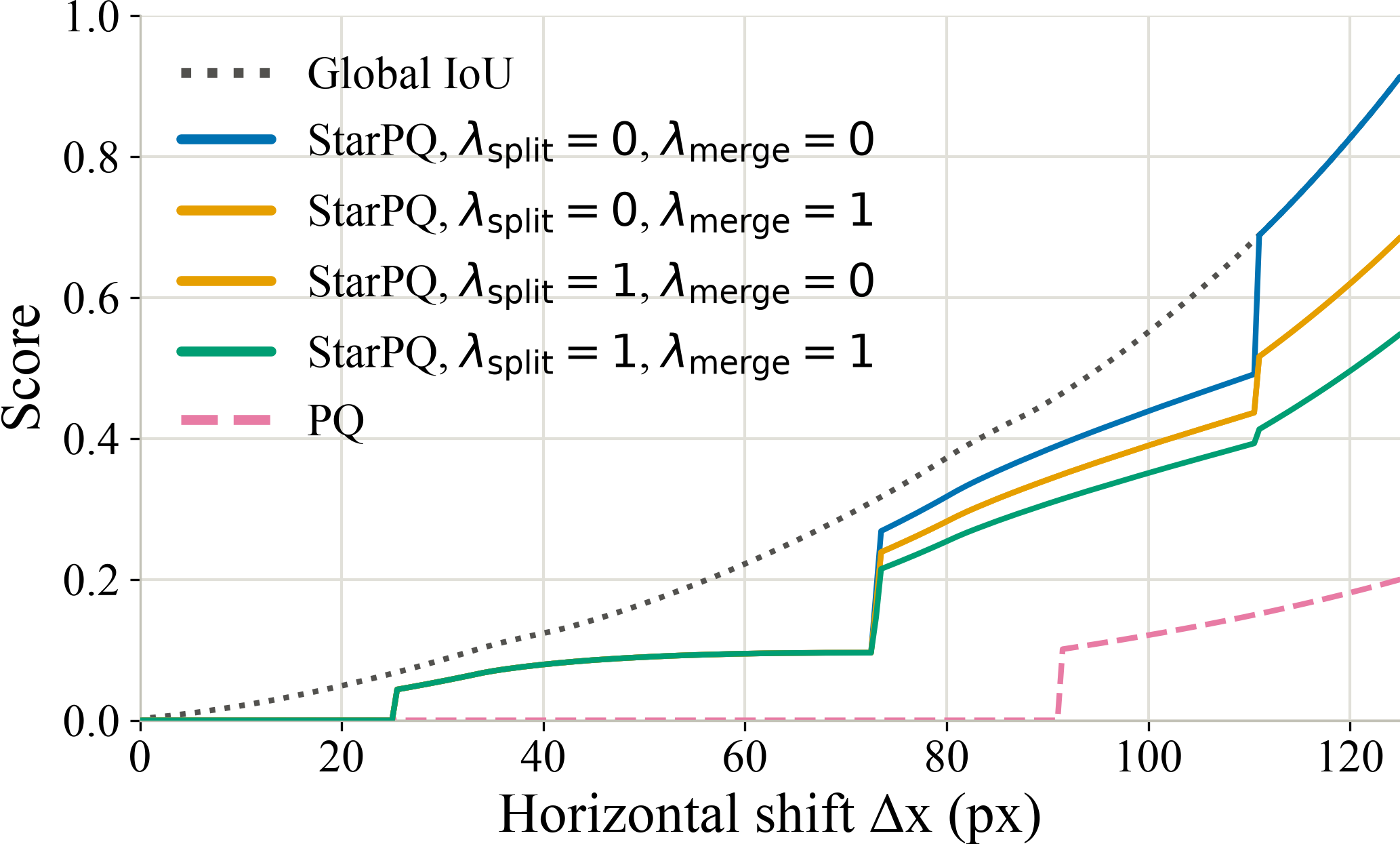}%
        \hspace{1em}%
        \includegraphics[height=4cm]{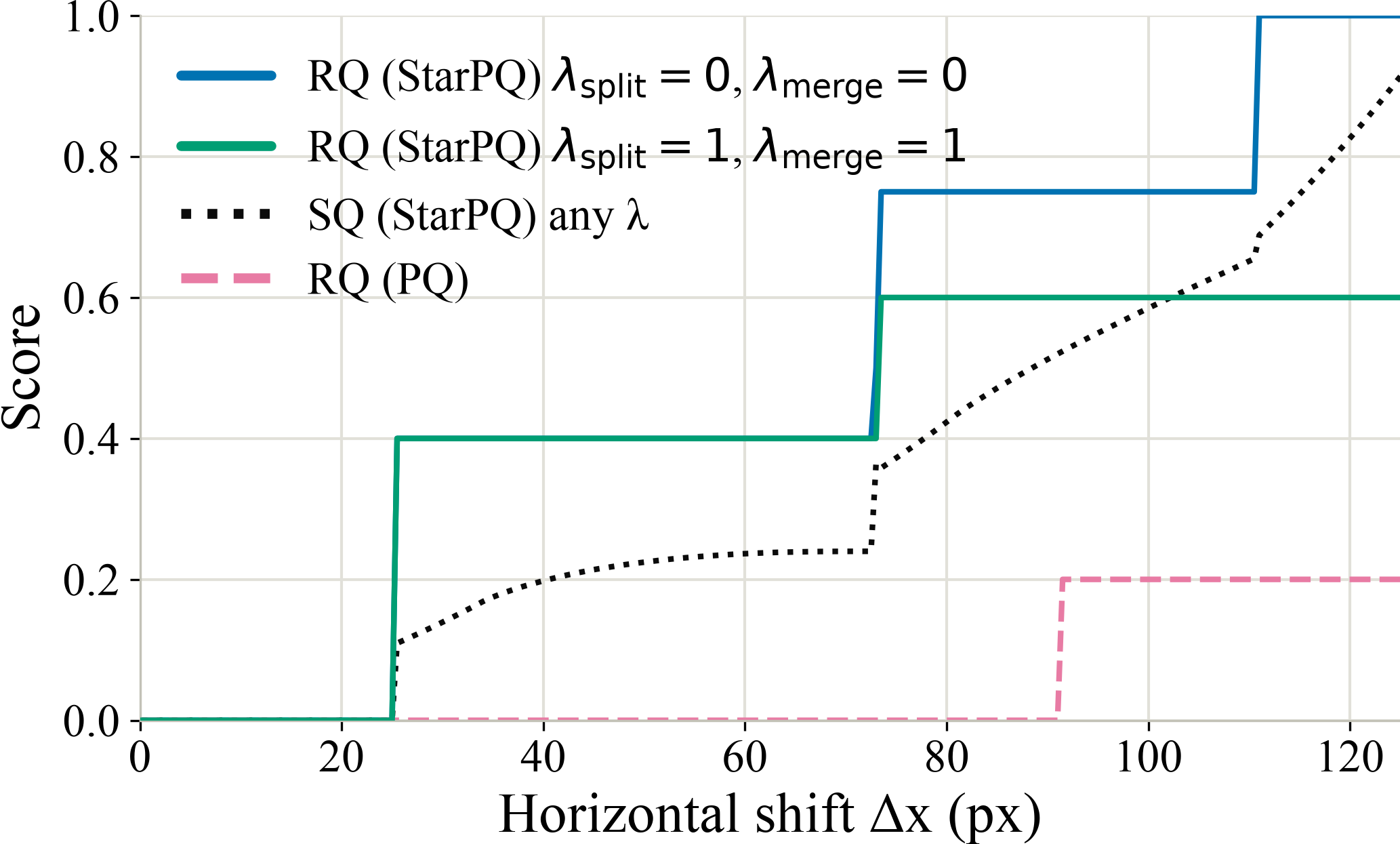}%
    }
    \caption{Response of IoU, PQ, and StarPQ (threshold $\tau=0$) to increasing overlap.
    \textbf{Left:} Initial configuration, in which the predictions (yellow) do not overlap the references (blue). 
    The bottom prediction circle and the top reference circle are fragmented into three instances.
    The arrow indicates the horizontal offset applied to the predictions, which is the x-axis of both plots.
    \textbf{Middle:} Global IoU, PQ, and StarPQ as a function of this offset. 
    PQ rises in a single step when the middle circles exceed the 0.5 IoU matching threshold.
    StarPQ shows this step earlier ($\sim$70px), as well as additional steps where fragments of the split segments satisfy the leaf criterion. 
    \textbf{Right:} Recognition Quality of PQ and StarPQ as a function of the offset.
    When predictions and references overlap at the end of the shift, PQ's RQ reports 0.2, since 2 out of 10 segments are matched.
    In contrast, StarPQ's RQ reports a perfect score when $\lambda_{\text{split}}=\lambda_{\text{merge}}=0$, because all segments are matched (one merge, one clean, and one split star).
    StarPQ's SQ curve is invariant to the configuration of $\lambda_{\text{split}}$ or $\lambda_{\text{merge}}$.
    }
    \label{fig:shift_score_figure}
\end{figure*}

%% file: body/4_experiments.tex
\section{Experiments} \label{sec:experiments}
In this section, we continue to compare \ac{StarPQ} against existing metrics and demonstrate the individual penalization of merges or splits, using a quantitative approach to provide the reader with a more intuitive understanding of \ac{StarPQ}'s behavior and characteristics.
We begin with a case study that isolates merge and split components, showing that \ac{StarPQ} detects them where \ac{PQ} cannot.
We next analyze instance-recovery methods from semantic maps and show how \ac{StarPQ} can be configured to account for their characteristic biases.
The same configurability helps when the ambiguity comes from the reference rather than the method, which we illustrate with an example of modeling expert disagreement.
Finally, we move from controlled settings to a benchmark across well-known datasets and models, where the decomposition into merge and split terms yields diagnostic information that reveals model characteristics a single score would hide.

\input{figures/instance_recovery/instance_recovery}
\input{tables/instance_decomposition}

\subsection{Splits and Merges}
To demonstrate \ac{StarPQ}'s ability to detect merge and split components, we conduct an experiment in which we linearly shift a prediction map over a reference map and thereby continuously increase their overlap (\cref{fig:shift_score_figure}).
The prediction map consists of three vertically stacked circles, where the bottom one is split into three separate instances.
The reference map is the prediction map flipped top-to-bottom, leading to the top prediction circle describing three references (merge star), the center circles mapping \ac{1:1}, and the bottom prediction split into three instances, jointly describing a single reference (split star).
\ac{PQ} exclusively detects the overlap between the whole center circles, as they are the only overlapping pair satisfying the criterion of an \ac{IoU} strictly greater than 0.5.
Therefore, \ac{PQ} increases in a single step when the center pair is counted as \ac{TP} ($\sim$90px).
In contrast, \ac{StarPQ}'s leaf criterion is generally more lenient by counting overlapping pairs earlier as \ac{TP}, since the required overlap is not tied to the \textit{union} of a pair, but only to the \textit{leaf's area}.
Hence, \ac{StarPQ} synchronously detects disjoint fragments of the bottom and top circles early ($\sim$25px) and also counts the center pair earlier ($\sim$70px) as a \ac{TP} compared to \ac{PQ}.
Note that this example of mirrored maps also illustrates \ac{StarPQ}'s symmetric handling of \ac{OS} and \ac{US}, as the $\lambda$ configurations that exclusively penalize one failure mode yield exactly the same result.

\subsection{Instance Recovery from Semantic Maps}
For many tasks that can be seen as instance segmentation problems, datasets with instance annotations do not exist.
If semantic maps exist, practitioners resort to recovering instance information from them.
Prior information about the shapes of the objects to be detected, how the dataset samples were obtained, and which failure modes can occur can be used to improve the quality of instance recovery from semantic maps.
\ac{StarPQ} can be used in such cases in two ways.
First, it can evaluate and analyze different instance approximation methods when both instance and semantic maps are available.
Second, it can configure leniency via $\lambda_{\text{split}}$ and $\lambda_{\text{merge}}$ towards systematically occurring failure modes of the instance recovery when only semantic maps exist.
To demonstrate these applications on real data, we conduct an experiment with the COCO dataset, where we fuse all instances for each thing class into a single semantic map and try to recover instance information.
\Cref{tab:instance_approximation} shows instance recovery across multiple classes via \ac{CC} on the left and by \ac{CC} and distance-transform watershed on the right.
In both tables, classes are sorted by the percentage of instances successfully recovered as clean stars by \ac{CC}, from high to low.
We show the three highest and lowest classes for \ac{CC}, separated by the dashed line.
\ac{StarPQ}'s star forest construction between the ground truth and each approximated instance map enables inspection of instance recovery by revealing failure modes.

\input{figures/annotations/expert_disagreement_figure}

\paragraph{Instance delineation priors.}
Methods that derive instances from semantic maps rely on a prior about how instances are separated, since semantic maps carry no instance information. 
\ac{CC} assumes that distinct instances are spatially disjoint, effectively meaning separated by at least one pixel. 
In natural images, which project a 3D scene onto a 2D grid, this logic fails in two ways: instances of the same class that touch or occlude one another are merged into a single component, and an instance partially covered by an instance of another class may be split into several.
\Cref{tab:instance_approximation} shows that \ac{CC} recovers clocks, microwaves, and baseball gloves most accurately, which aligns with our intuition:
There is little reason to place multiple clocks next to each other, and it is important that clocks are not covered and can be clearly viewed from many perspectives (\cref{fig:instance_recovery}).
\ac{CC} performs worst when attempting to recover skis, sheep, and donuts:
Skis are an outlier in the COCO dataset, as a pair is considered a single instance, leading to splits.
Mutual occlusion is expected for herd animals like sheep, leading to merges.
Similarly, donuts lined up next to each other may merge into a single instance.
Such class-specific priors allow \ac{StarPQ} to be configured against the known bias of the instance approximation.
For classes where \ac{CC} predominantly merges instances (sheep, donuts), setting $\lambda_{\text{split}}=0$ and $\lambda_{\text{merge}}=1$ forgives splits predicted by a model that correctly separates instances that the approximation merged, while still penalizing merges. 
For classes where \ac{CC} predominantly splits (skis), the opposite setting applies.
Refining \ac{CC} with a distance-transform watershed, which further splits each component where it is thinnest, can improve instance recovery in some classes, at the cost of additional splits in others.
Distance-transform watershed adds a geometric prior about shape by assuming that each instance has a single dominant distance maximum and that touching instances meet where the component is thinnest, which holds for disk-like instances of comparable size such as donuts in a tray, but not for elongated ones such as skis (\cref{tab:instance_approximation}).

\input{figures/annotations/expert_disagreement_table}

\subsection{Instance-Wise Penalization}
So far, $\lambda_{\text{split}}$ and $\lambda_{\text{merge}}$ were set once for a whole instance map or class and thus applied uniformly to all of its instances.
When multiple expert instance annotations exist, and experts disagree on splitting or merging specific instances, \ac{StarPQ} may be configured at an \textit{instance level}:
\ac{StarPQ}'s star forest construction from multiple expert annotations, each against the compiled reference (\eg by majority vote), allows detection of \textit{expert delineation disagreement} by revealing which instances in the reference are split or merged by one or more experts.
If sufficiently many experts split or merge particular instances, the reference map can be supplemented by one set of instance labels that may legitimately be merged and another that accounts for valid splits. 
These sets can be used as additional information for the \ac{StarPQ} evaluation by forgiving models that merge or split instances that lie in each set, respectively, thereby allowing the expression of expert uncertainty about instance delineation in the evaluation.
\Cref{tab:expert_annotations} applies this to the example of \cref{fig:expert_annotations}: with $\lambda_{\text{split}} = 0$ for s and $\lambda_{\text{merge}} = 0$ for $\text{m}_1$ and $\text{m}_2$, the split of expert 2 and the merge of expert 3 are forgiven and all three experts reach a perfect \ac{RQ}, whereas \ac{PQ} and \ac{StarPQ} with $\lambda = 1$ penalize both.
An interesting future direction is exploring how this information may serve as a signal during model training.

\input{tables/coco_benchmark}

\paragraph{Real lambda values.}
By definition, $\lambda$ values are either 0 or 1, and the resulting weighted \ac{FP}/\ac{FN} counts are integers.
$\FP^{w}$ counts all unmatched predictions \ac{FP} and additionally all extra edges contributing to split stars for $\lambda_{\text{split}} = 1$.
$\FN^{w}$ is defined analogously with $\lambda_{\text{merge}}$ and merge stars.
This restriction makes the counts interpretable because segments are counted as whole units, whereas counting fractions of segments would introduce ambiguity.
However, from an implementation standpoint, real values are possible and allow for more fine-grained adjustment of the penalization of merges and splits.
We argue that this reduces interpretability because $\FP^{w}$ and $\FN^{w}$ no longer correspond to discrete, identifiable instances.
However, in the context of \textit{instance-level} configuration, we may use the fraction of experts that do not split or merge an instance to define instance-level $\lambda$ values.
In the example of \cref{fig:expert_annotations}, this would translate to configuring $\lambda_{\text{merge}} = \frac{2}{3}$ for both instances merged by expert 3 ($\text{m}_1$, $\text{m}_2$) and $\lambda_{\text{split}} = \frac{2}{3}$ for the instance split by expert 2 (s), since two of the three annotators did not merge or split the corresponding instances, respectively.
A split star is charged with the $\lambda_{\text{split}}$ of its reference instance, and a merge star charges each of its reference instances with its own $\lambda_{\text{merge}}$, except the one with the smallest value.
A merge is therefore only forgiven among instances that all carry a low $\lambda_{\text{merge}}$.
Although $\lambda > 1$ is technically possible, we do not consider it meaningful, as it would count each extra segment more than once in the \ac{RQ} term and thereby contradict the at-most-once counting principle.

\subsection{Model Baselines}
Finally, we show across the COCO~\citep{linMicrosoftCOCOCommon2014} (\cref{tab:coco_things_pq_vs_starpq}) and Cityscapes~\citep{cordts2016cityscapes} (\cref{tab:cityscapes_things_pq_vs_starpq}) datasets how well-known baseline models~\citep{kirillovPanopticFeaturePyramid2019, chengMaskedattentionMaskTransformer2022, NEURIPS2021_950a4152} score across different configurations.
For \ac{StarPQ}, we report configurations where only splits or only merges are penalized.
We set the star threshold $\tau=0.5$ for \ac{StarPQ} to match \ac{PQ}'s matching criterion to increase comparability.
Additionally, we report how many reference instances were recovered cleanly, as splits, or as merges.
In both datasets, merges occur more frequently than splits, while the difference is greater in Cityscapes.
Hence, all models are more prone to \ac{US} than to \ac{OS}. 

\input{tables/cityscapes_benchmark}

%% file: figures/instance_recovery/instance_recovery.tex
\begin{figure*}[t]
    \includegraphics[width=\textwidth]{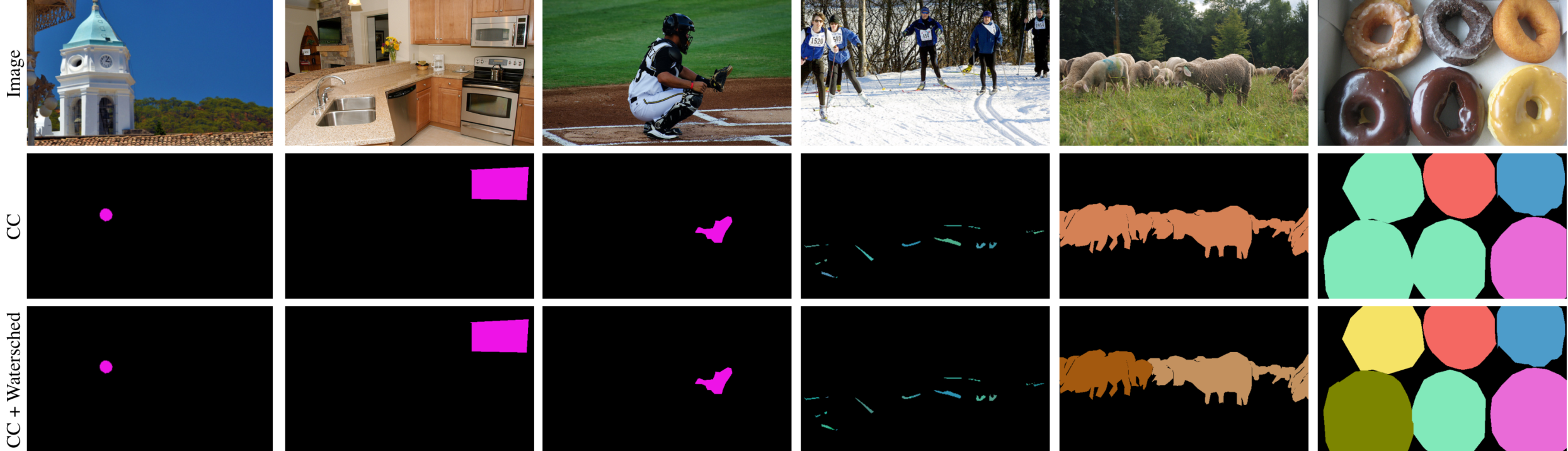}
    \caption{Examples of classes where CC recovered most of the instances cleanly (clock, microwave, and baseball glove) and where the fewest instances could be recovered 1:1 (skis, sheep, and donut).
    The first row shows images, the second row instance maps recovered by CC, and the third row instance maps recovered by CC + watershed.
    Isolated instances can often be cleanly recovered, whereas mutually occluding or touching instances, such as herd animals or donuts, are merged by CC. Watershed can additionally separate compact, disk-like shapes such as donuts.}
    \label{fig:instance_recovery}
\end{figure*}

%% file: tables/instance_decomposition.tex
\begin{table*}[t]
    \centering
    \setlength{\tabcolsep}{4pt}
    \caption{Instance approximation of COCO panoptic val2017 thing classes by (a) connected components and (b) connected components, split by a distance-transform watershed.
    Clean, split, and merge: percentage of the reference instances recovered
    one-to-one, over-segmented, or absorbed into a merge. 
    The remaining reference instances are unmatched.
    StarPQ at $(\lambda_{\text{split}}, \lambda_{\text{merge}})$ as labeled, and all $\times 100$.
    Rows: the 3 classes with the highest and the lowest clean percentage in (a), sorted in descending order, in the same order in (b). 
    Touching or mutually occluding instances of classes like sheep or donuts are systematically merged by CC. 
    When such a CC map serves as the reference, a model that correctly
    separates these instances produces splits, which the configuration
    lenient towards oversegmentation
    $(\lambda_{\text{split}}=0, \lambda_{\text{merge}}=1)$ forgives.
    }
    \label{tab:instance_approximation}
    \begin{tabular}[t]{@{}lrrrrrr@{}}
        \toprule
        \multicolumn{7}{@{}l}{(a) Connected components (CC)} \\
        \addlinespace
        & \multicolumn{3}{c}{Instances \%} & \multicolumn{2}{c}{StarPQ} & \\
        \cmidrule(lr){2-4} \cmidrule(lr){5-6}
        Class & Clean & Split & Merge & $(1, 0)$ & $(0, 1)$ & PQ \\
        \midrule
        clock & 98.9 & 0.4 & 0.7 & 99.1 & 99.8 & 98.7 \\
        microwave & 98.1 & 1.9 & 0.0 & 99.1 & 100.0 & 99.0 \\
        baseball glove & 97.3 & 0.7 & 2.0 & 99.7 & 99.3 & 98.5 \\
        \noalign{\vskip\aboverulesep}\hdashline\noalign{\vskip\belowrulesep}
        skis & 24.9 & 73.0 & 0.0 & 62.7 & 97.6 & 42.8 \\
        sheep & 23.4 & 3.1 & 73.2 & 86.3 & 57.2 & 39.1 \\
        donut & 22.2 & 0.0 & 77.8 & 99.2 & 57.6 & 38.6 \\
        \midrule
        Mean & 63.7 & 11.3 & 24.8 & 87.9 & 88.6 & 73.2 \\
        \bottomrule
    \end{tabular}
    \hfill
    \begin{tabular}[t]{@{}lrrrrrr@{}}
        \toprule
        \multicolumn{7}{@{}l}{(b) CC + distance-transform watershed} \\
        \addlinespace
        & \multicolumn{3}{c}{Instances \%} & \multicolumn{2}{c}{StarPQ} & \\
        \cmidrule(lr){2-4} \cmidrule(lr){5-6}
        Class & Clean & Split & Merge & $(1, 0)$ & $(0, 1)$ & PQ \\
        \midrule
        clock & 88.8 & 11.2 & 0.0 & 92.7 & 100.0 & 87.4 \\
        microwave & 92.6 & 7.4 & 0.0 & 96.4 & 100.0 & 94.0 \\
        baseball glove & 91.9 & 8.1 & 0.0 & 95.5 & 100.0 & 91.8 \\
        \noalign{\vskip\aboverulesep}\hdashline\noalign{\vskip\belowrulesep}
        skis & 14.9 & 83.8 & 0.0 & 40.1 & 98.1 & 20.4 \\
        sheep & 26.8 & 6.2 & 66.7 & 82.3 & 65.8 & 44.1 \\
        donut & 38.9 & 1.9 & 58.3 & 95.4 & 72.0 & 57.1 \\
        \midrule
        Mean & 59.0 & 21.0 & 19.4 & 79.7 & 90.8 & 65.5 \\
        \bottomrule
    \end{tabular}
\end{table*}

%% file: figures/annotations/expert_disagreement_figure.tex
\newlength{\annotationImageSize}
\newlength{\annotationGridGap}
\begin{figure}[!t]
    \centering
    \setlength{\annotationGridGap}{2mm}
    \setlength{\annotationImageSize}{\dimexpr(\linewidth - \annotationGridGap) / 2\relax}
    \begin{subfigure}[t]{\annotationImageSize}
        \centering
        \includegraphics[width=\linewidth]{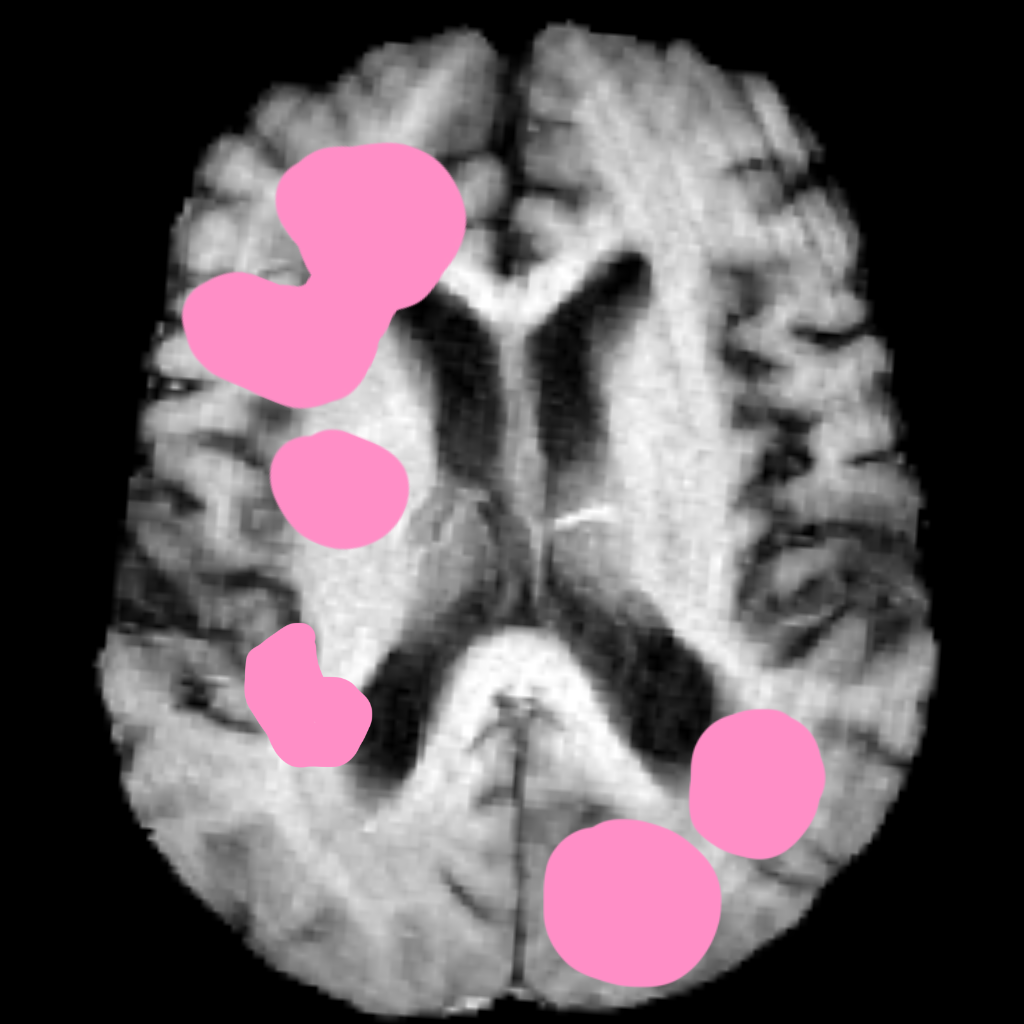}
        \caption{Expert 1}
        \label{fig:expert_annotations:expert_1}
    \end{subfigure}\hspace{\annotationGridGap}%
    \begin{subfigure}[t]{\annotationImageSize}
        \centering
        \includegraphics[width=\linewidth]{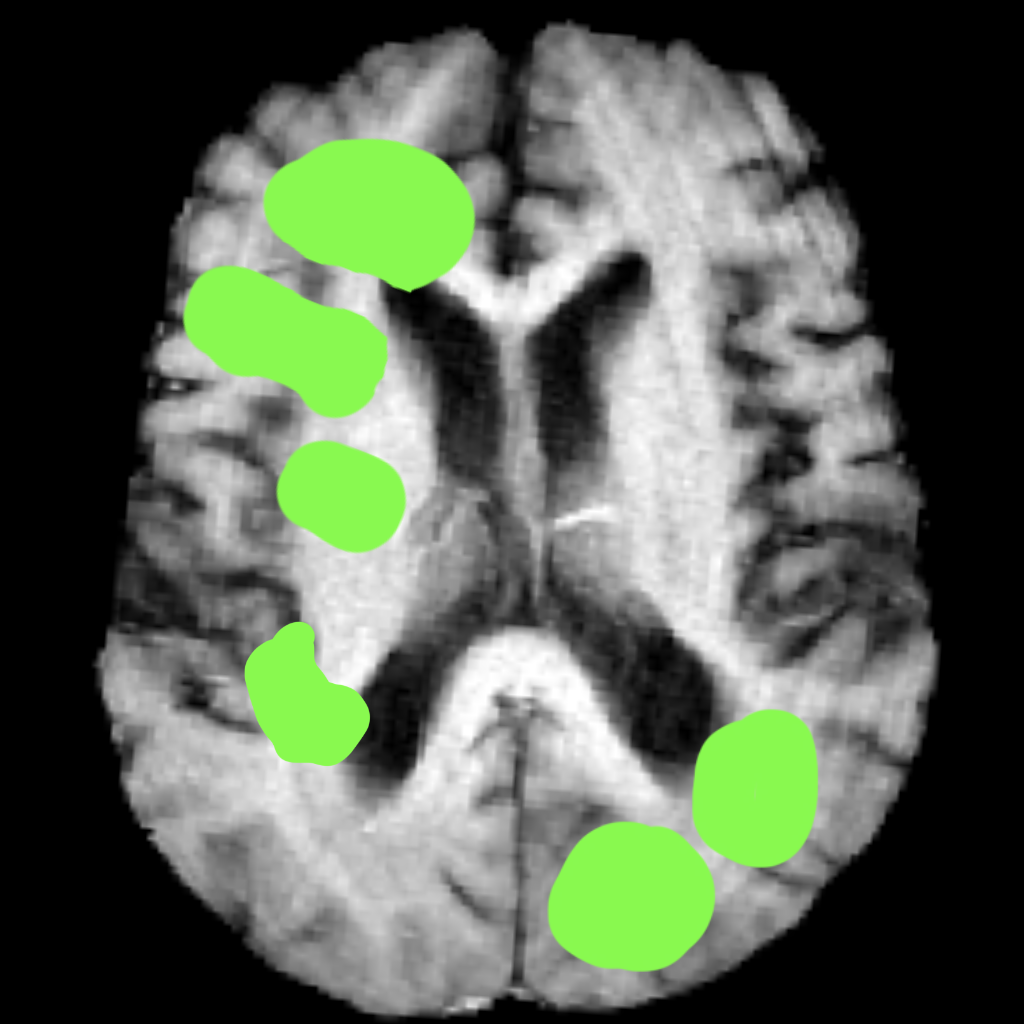}
        \caption{Expert 2}
        \label{fig:expert_annotations:expert_2}
    \end{subfigure}

    \medskip
    \begin{subfigure}[t]{\annotationImageSize}
        \centering
        \includegraphics[width=\linewidth]{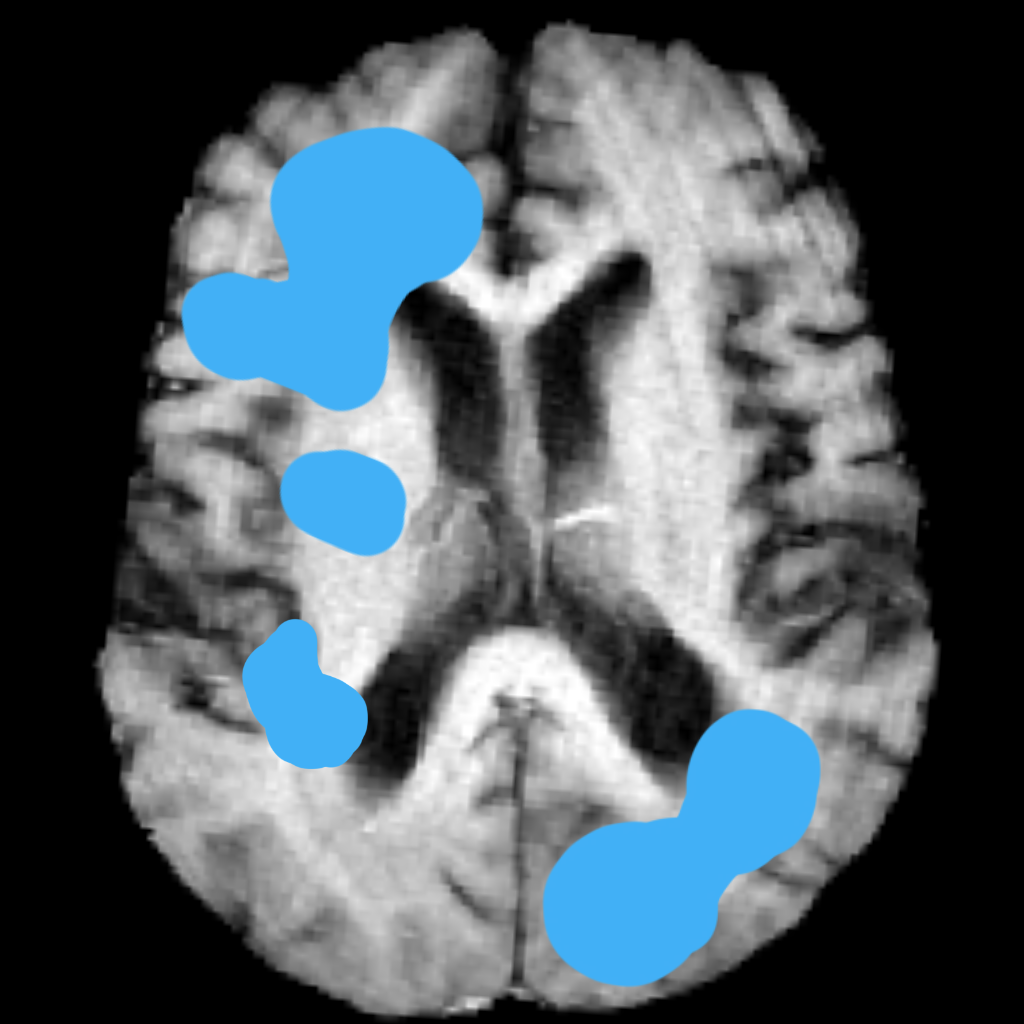}
        \caption{Expert 3}
        \label{fig:expert_annotations:expert_3}
    \end{subfigure}\hspace{\annotationGridGap}%
    \begin{subfigure}[t]{\annotationImageSize}
        \centering
        \includegraphics[width=\linewidth]{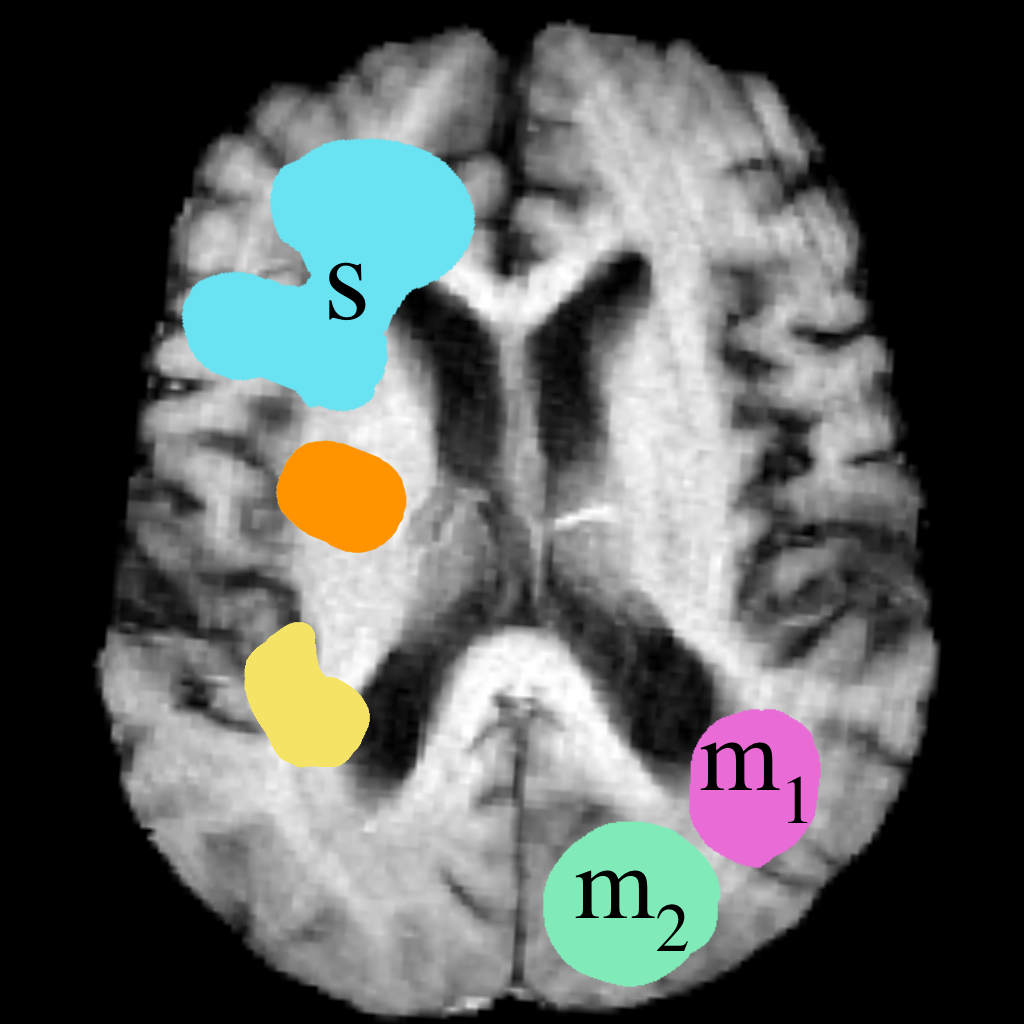}
        \caption{Majority vote}
        \label{fig:expert_annotations:majority_vote}
    \end{subfigure}

    \caption{Illustrative example of MRI metastases annotations of the same slice by three experts \textbf{(a)--(c)} and a majority vote \textbf{(d)}, acting as the reference, in which a pixel is labeled if at least two of the three experts labeled it.
    The majority vote coincides with the instance delineation of expert 1.
    Expert 2 splits the top instance (s) into two instances, and expert 3 merges the two bottom instances ($\text{m}_1$ and $\text{m}_2$) into one.}
    \label{fig:expert_annotations}
\end{figure}

%% file: figures/annotations/expert_disagreement_table.tex
\begin{table}[t]
    \centering
    \caption{Scores of each expert in \cref{fig:expert_annotations} against the majority vote.
    Expert 1 scores a perfect RQ across all configurations.
    PQ's RQ and StarPQ's RQ with $\lambda = 1$ penalize both the split of expert 2 and the merge of expert 3.
    We construct the star forest for each expert annotation map against the majority vote and use the detected splits and merges to model leniency via $\lambda_{\text{split}}$ and $\lambda_{\text{merge}}$ at an \textit{instance level} for StarPQ, here with $\lambda_{\text{split}} = 0$ for s and $\lambda_{\text{merge}} = 0$ for $\text{m}_1$ and $\text{m}_2$.
    The instance-level configuration (instance $\lambda$) lets each expert score a perfect RQ.
    }    \label{tab:expert_annotations}
    \small
    \setlength{\tabcolsep}{3pt}
    \begin{tabular*}{\linewidth}{@{\extracolsep{\fill}}lcccccc@{}}
        \toprule
        & \multicolumn{2}{c}{PQ}
        & \multicolumn{2}{c}{\begin{tabular}[b]{@{}c@{}}StarPQ\\($\lambda=1$)\end{tabular}}
        & \multicolumn{2}{c}{\begin{tabular}[b]{@{}c@{}}StarPQ\\(instance $\lambda$)\end{tabular}} \\
        \cmidrule(lr){2-3} \cmidrule(lr){4-5} \cmidrule(lr){6-7}
        & Score & RQ & Score & RQ & Score & RQ \\
        \midrule
        (a) Expert 1 & 91.1 & 100.0 & 91.1 & 100.0 & 91.1 & 100.0 \\
        (b) Expert 2 & 67.9 &  72.7 & 83.8 &  90.9 & 92.2 & 100.0 \\
        (c) Expert 3 & 74.1 &  88.9 & 82.8 &  88.9 & 93.1 & 100.0 \\
        \bottomrule
    \end{tabular*}
\end{table}

%% file: tables/coco_benchmark.tex
\begin{table}[t]
    \centering
    \setlength{\tabcolsep}{4pt}
    \caption{Models on the 80 thing classes of COCO panoptic val2017, sorted by PQ from high to low.
    M2F: Mask2Former; MF: MaskFormer; FPN: Panoptic FPN; RN-50: ResNet-50; Swin-B and Swin-L: Swin base and large.
    Clean, split and merge: percentage of the reference instances recovered One-to-One, over-segmented, or absorbed into a merge. StarPQ at $(\lambda_{\mathrm{split}}, \lambda_{\mathrm{merge}})$ as labeled and all $\times 100$.
    Every column is the mean over the classes.}
    \label{tab:coco_things_pq_vs_starpq}
    \begin{tabular*}{\columnwidth}{@{\extracolsep{\fill}}lrrrrrr@{}}
        \toprule
        & \multicolumn{3}{c}{Instances \%} & \multicolumn{2}{c}{StarPQ} & \\
        \cmidrule(lr){2-4} \cmidrule(lr){5-6}
        Model & Clean & Split & Merge & $(1, 0)$ & $(0, 1)$ & PQ \\
        \midrule
        M2F Swin-L & 70.2 & 1.6 & 5.1 & 65.0 & 64.8 & 62.6 \\
        M2F Swin-B & 66.2 & 1.8 & 5.3 & 62.0 & 61.9 & 59.6 \\
        MF Swin-B & 64.8 & 2.6 & 4.7 & 57.9 & 58.1 & 55.4 \\
        FPN RN-50 & 49.2 & 0.9 & 4.0 & 50.1 & 49.8 & 48.3 \\
        \bottomrule
    \end{tabular*}
\end{table}

%% file: tables/cityscapes_benchmark.tex
\begin{table}[t]
    \centering
    \setlength{\tabcolsep}{4pt}
    \caption{Models on the 8 thing classes of Cityscapes panoptic val, sorted by PQ from high to low.
    M2F: Mask2Former; Swin-T, Swin-S, Swin-B and Swin-L: Swin tiny, small, base and large.
    Clean, split and merge: percentage of the reference instances recovered One-to-One, over-segmented, or absorbed into a merge. StarPQ at $(\lambda_{\mathrm{split}}, \lambda_{\mathrm{merge}})$ as labeled and all $\times 100$.
    Every column is the mean over the classes.}
    \label{tab:cityscapes_things_pq_vs_starpq}
    \begin{tabular*}{\columnwidth}{@{\extracolsep{\fill}}lrrrrrr@{}}
        \toprule
        & \multicolumn{3}{c}{Instances \%} & \multicolumn{2}{c}{StarPQ} & \\
        \cmidrule(lr){2-4} \cmidrule(lr){5-6}
        Model & Clean & Split & Merge & $(1, 0)$ & $(0, 1)$ & PQ \\
        \midrule
        M2F Swin-L & 62.1 & 0.5 & 9.2 & 61.3 & 59.9 & 58.2 \\
        M2F Swin-B & 62.1 & 0.6 & 9.0 & 60.6 & 59.3 & 57.6 \\
        M2F Swin-S & 59.8 & 0.9 & 9.9 & 59.2 & 57.9 & 55.7 \\
        M2F Swin-T & 60.5 & 0.7 & 8.6 & 58.3 & 57.1 & 55.1 \\
        \bottomrule
    \end{tabular*}
\end{table}

%% file: body/5_conclusion.tex
\section{Limitations and Conclusion} \label{sec:conclusion}
We present \ac{StarPQ}, a novel metric in the spirit of \acl{PQ} that does not rely on a strict \acl{1:1} assignment but instead decomposes \aclp{TP} into three categories measuring split, merge, and clean components.
We believe that this metric not only enables more precise evaluation of instance segmentation models but also serves as a diagnostic tool for revealing instance delineation failure modes.
A limitation of \ac{StarPQ} is that its star forest is constructed by a heuristic and is therefore not guaranteed to be optimal, i.e., to maximize the sum of star \acp{IoU}.
The heuristic can deviate from the optimum when pointers form chains.
Since chains are rare in practice (\cref{sec:method}), we accept this trade-off in favor of a matching that is simple and fast to compute.
We further encourage the community to consider prior information arising from dataset acquisition, instance approximation, ground-truth composition methods, and requirements for downstream applications that may affect the evaluation of instance segmentation pipelines.

%% file: body/6_llm_usage_statement.tex
\section{LLM Usage Statement} \label{sec:llm_usage_statement}
The elephant image in \cref{fig:examples} was generated with \textit{Gemini 3 Pro Image}. 
Its semantic and instance maps are predictions of Mask2Former Swin-L. \textit{Claude Opus 5.5} was used to generate parts of the code for the experiments and the metric implementation. 
The authors wrote most of the manuscript's text; \textit{Claude Opus 5.5} refined some sentences for readability, detected spelling and grammar errors, and provided high-level feedback during writing.

%% file: body/7_acknowledgements.tex
\section{Acknowledgments} \label{sec:acknowledgments}
This work was supported by the Gemeinnützige Hertie-Stiftung.
The authors thank the International Max Planck Research School for Intelligent Systems (IMPRS-IS) for supporting EG.
FK appreciates the support of Anna Valentina Lioba Eleonora Claire Javid Mamasani.
The authors declare no conflict of interest.

%% file: body/8_appendix.tex
\twocolumn[{%
  \section{Appendix}\label{sec:appendix}
  \captionsetup{type=figure}%
  \centering
  \resizebox{\textwidth}{!}{%
    \includegraphics[height=4cm]{figures/overlap_shift/shift_score_image.png}%
    \hspace{1em}%
    \includegraphics[height=4cm]{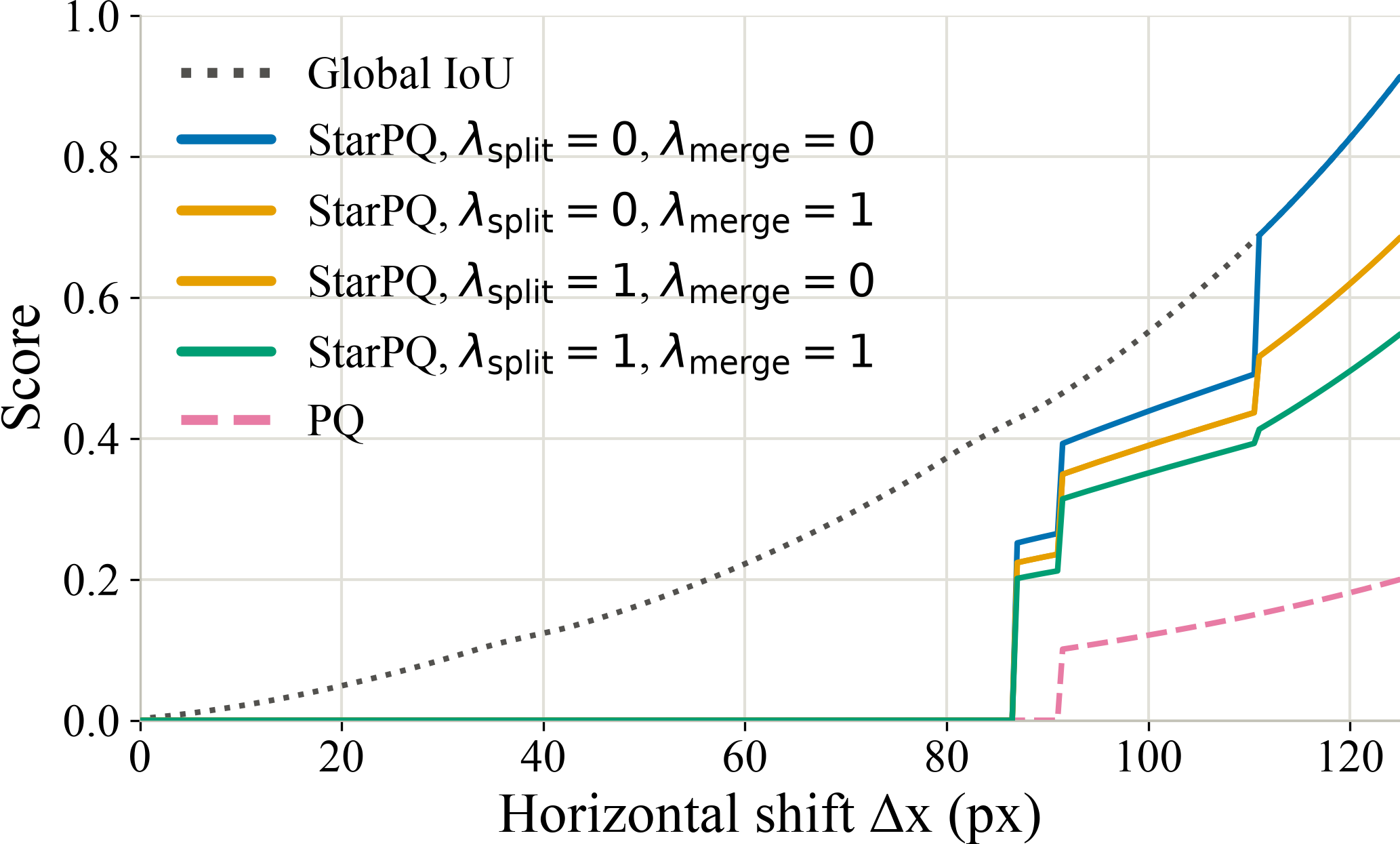}%
    \hspace{1em}%
    \includegraphics[height=4cm]{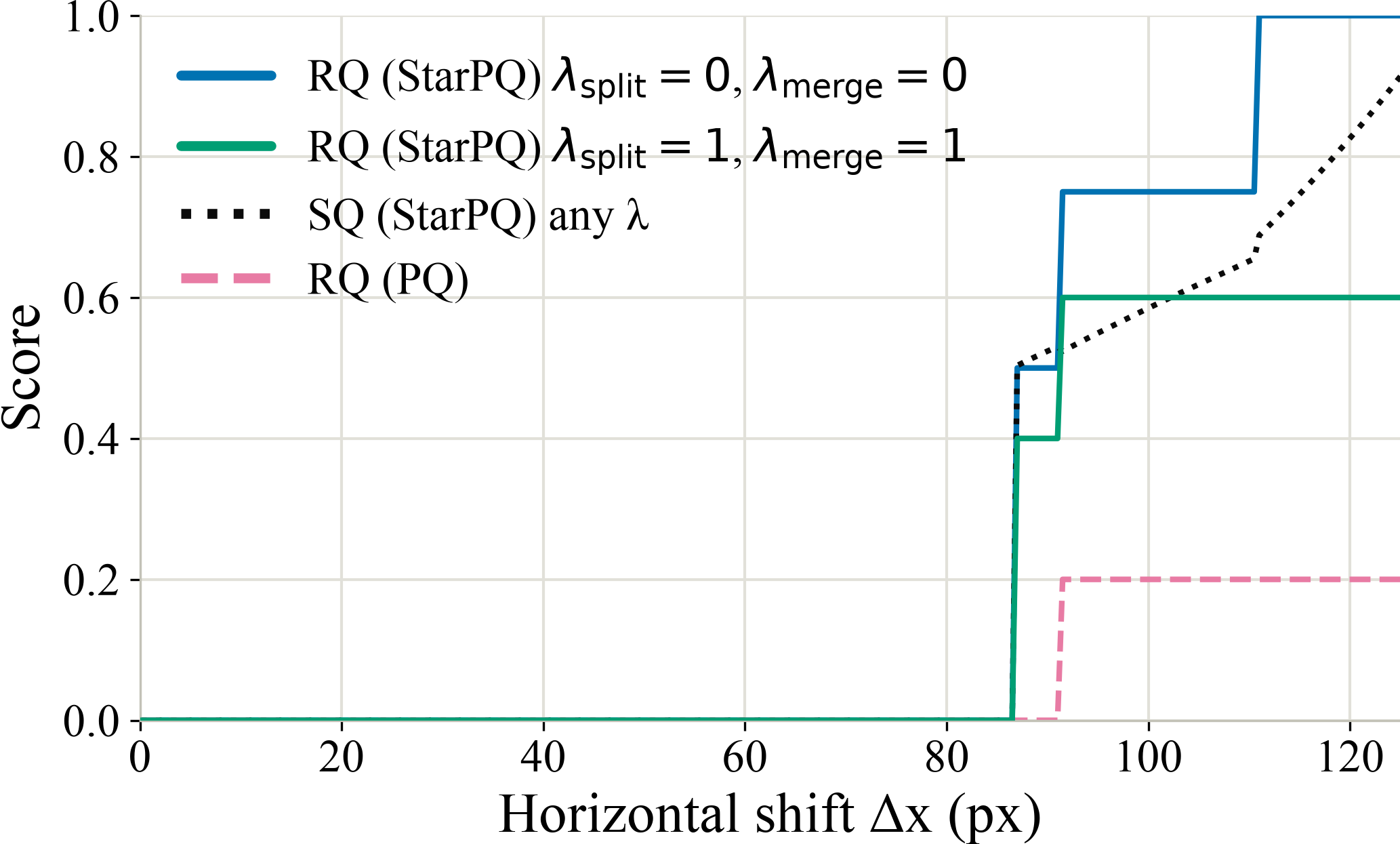}%
  }
  \caption{Response of IoU, PQ, and StarPQ to increasing overlap, with star threshold $\tau=0.5$.
    The setup matches \cref{fig:shift_score_figure}, except that $\tau=0.5$ instead of $\tau=0$, to match PQ's threshold.
    StarPQ still rises earlier than PQ, because the overlapping pairs satisfy the leaf criterion before they satisfy PQ's IoU criterion.}%
  \label{fig:shift_score_figure_0_5}
  \vspace{1em}
}]

%% file: config/rules.tex
\section{\LaTeX{} evangelium} \label{sec:evangelium}
This section appears only in debug mode.
Each example shows the source command and, where useful, its output.

\subsection*{Labels and references}
\begin{itemize}
    \item Label everything you reference, with a type prefix: \verb|\label{sec:...}|, \verb|\label{fig:...}|, \verb|\label{tab:...}|, \verb|\label{eq:...}|.
    \item Place a label directly after its \verb|\section| or \verb|\caption|, or inside its equation environment.
    A label before the caption picks up the wrong number.
    \item Reference with \verb|\cref{sec:evangelium}| (\cref{sec:evangelium}), or with \verb|\Cref{...}| at the start of a sentence (\Cref{sec:evangelium}).
    \item Reference several targets at once with \verb|\cref{fig:a,fig:b}|; cleveref sorts and groups them.
\end{itemize}

\subsection*{Figures and tables}
\begin{itemize}
    \item Put each figure or table in its own file, \texttt{figures/<name>.tex} or \texttt{tables/<name>.tex}, and \verb|\input| it where it belongs.
    Moving a float during final layout then means moving one line.
    \item Store plots and diagrams in \texttt{figures/} as vector PDF.
    Store images such as scans and segmentation overlays as PNG, never JPEG: compression artifacts blur label boundaries.
    \item Use color-blind-safe colors: the palette of \href{https://www.nature.com/articles/nmeth.1618}{Wong (2011)} for categories, and perceptually uniform colormaps such as viridis or cividis for continuous values.
    \item Never encode meaning by color alone; add markers, line styles, or labels.
    \item Use booktabs rules (\verb|\toprule|, \verb|\midrule|, \verb|\bottomrule|) and no vertical lines in tables.
    \item Start from the templates \verb|figures/example_figure.tex| (\cref{fig:example_figure}) and \verb|tables/example_table.tex| (\cref{tab:example_table}).
    \item Place captions above tables and below figures.
    \item Use \verb|\centering| inside floats, not the \verb|center| environment, which adds extra vertical space.
    \item Size graphics relative to the layout, e.g.\ \verb|width=0.6\linewidth|, rather than in absolute units.
    \item In result tables, give every number the same number of decimal places, mark the best result in bold, and state the metric direction with $\uparrow$ or $\downarrow$.
\end{itemize}

\input{figures/example_figure.tex}
\input{tables/example_table.tex}

\subsection*{Citations}
\begin{itemize}
    \item Add entries to \texttt{references.bib}, keyed as first author, year, and first title word, e.g.\ \texttt{cordts2016cityscapes}.
    \item Use \verb|\citep{...}| for parenthetical citations (\citep{cordts2016cityscapes}) and \verb|\citet{...}| when the authors are part of the sentence (\citet{cordts2016cityscapes}).
\end{itemize}

\subsection*{Acronyms}
\acresetall %
\begin{itemize}
    \item Define acronyms in \texttt{config/acronyms.tex}, never inline in the text.
    \item Use \verb|\ac{ROI}| everywhere; the first use expands, later uses do not.
    First use: \ac{ROI}; later: \ac{ROI}.
    \item Plural: \verb|\acp{ROI}| (\acp{ROI}).
    Long form only: \verb|\acl{ROI}| (\acl{ROI}), plural \verb|\aclp{ROI}| (\aclp{ROI}).
    \item Indefinite article by pronunciation: \verb|\iac{FP}| (\iac{FP}); \verb|\Iac{FP}| at the start of a sentence.
    \item In captions, subcaptions and table cells, write acronyms as plain text: floats move, so the first-use expansion of \verb|\ac| may land out of reading order.
    \item In section titles, write acronyms as plain text: titles also go into the PDF bookmarks.
    \item In math, use the macros of \texttt{config/commands.tex} (\verb|\TP|, \verb|\FP|, \verb|\FN|, \verb|\IoU|, \verb|\IoL|, \verb|\PQ|, \verb|\SQ|, \verb|\RQ|, \verb|\StarPQ|) and write word subscripts as \verb|\text{...}|, e.g.\ \verb|\lambda_{\text{split}}|.
\end{itemize}

\subsection*{Drafting: TODOs and debug information}
The switch \verb|\debugtrue| or \verb|\debugfalse| at the top of \texttt{main.tex} controls everything below, including whether this section appears.
\begin{itemize}
    \item Notes about the paper: \verb|\todo{...}| renders as \todo{check out this cool command}.
    \item Unfinished paper text: wrap it in \verb|\begin{todoblock}{reason}| \dots\ \verb|\end{todoblock}|.
    In final mode the text is typeset normally, and each remaining block logs a warning.
    \item Author-only information: \verb|\debug{...}| renders as \debug{debug information}, and \verb|debugblock| does the same for whole blocks.
    Both vanish in final mode.
    \item Before submitting, compile with \verb|\debugfalse| and search the log for ``Unresolved TODO''; it must find nothing.
    \item Verbatim (\verb|\verb|, listings) does not work inside these commands or blocks.
    \item Floats inside a block move away from its markers; mark them with \verb|\todo| or \verb|\debug| in the caption.
\end{itemize}

\begin{todoblock}{example reason}
This paragraph is draft text that still needs work.
\end{todoblock}

\subsection*{Writing and source style}
\begin{itemize}
    \item Write in the present tense.
    \item Write one sentence per line, so git diffs show changed sentences instead of reflowed paragraphs.
    \item Format and lint the source with \texttt{latexindent} (it has a one-sentence-per-line setting) and \texttt{chktex}.
    \item Use a spellchecker such as Grammarly or LanguageTool.
\end{itemize}

%% file: figures/example_figure.tex
\begin{figure}[t]
    \centering
    \includegraphics[width=0.6\linewidth]{example-image}
    \caption{Example figure.
    Replace \texttt{example-image} with a graphic stored in \texttt{figures/}, e.g.\ \texttt{figures/overview.pdf}.}
    \label{fig:example_figure}
\end{figure}

%% file: tables/example_table.tex
\begin{table}[t]
    \caption{Example table with placeholder values.
    Best results in bold; $\uparrow$: higher is better.}
    \label{tab:example_table}
    \centering
    \begin{tabular}{lrrr}
        \toprule
        Method     & PQ $\uparrow$ & SQ $\uparrow$ & RQ $\uparrow$ \\
        \midrule
        Baseline A & 61.2          & 80.4          & 74.1          \\
        Baseline B & 63.5          & 81.0          & 76.3          \\
        Ours       & \textbf{66.8} & \textbf{82.7} & \textbf{79.5} \\
        \bottomrule
    \end{tabular}
\end{table}